\documentclass[sigconf]{acmart}
\AtBeginDocument{%
  }

\usepackage{booktabs}
\usepackage{graphicx}
\usepackage{pifont}
\usepackage{transparent}
\usepackage{threeparttable}
\usepackage{multirow}
\usepackage[table]{xcolor}
\usepackage{enumitem}

\renewcommand\footnotetextcopyrightpermission[1]{}
\setcopyright{none}
\copyrightyear{}
\acmYear{}
\acmDOI{}
\acmConference{}
\acmISBN{}

\begin{document}

\title{Pneuma: A Large-Scale Real-World Time Series Benchmark for Commercial Aircraft Predictive Maintenance}

\author{Jiafeng Lin}
\authornote{Authors contributed equally to this research.}
\email{lin-jf21@mails.tsinghua.edu.cn}
\orcid{0009-0004-5299-9992}
\affiliation{%
  \institution{School of Software, BNRist, \\ Tsinghua University}
  \city{Beijing}
  \country{China}
}

\author{Yutian Zhang}
\authornotemark[1]
\email{zhangyt2002@gmail.com}
\affiliation{%
  \institution{School of Software, BNRist, \\ Tsinghua University}
  \city{Beijing}
  \country{China}
}

\author{Huan Zhang}
\authornotemark[1]
\email{zhanghuan\_a@csair.com}
\affiliation{%
  \institution{China Southern Airlines}
  \city{Guangzhou}
  \country{China}
}

\author{Yuxuan Wang}
\email{wangyuxu22@mails.tsinghua.edu.cn}
\orcid{0000-0002-4899-4716}
\affiliation{%
  \institution{School of Software, BNRist, \\ Tsinghua University}
  \city{Beijing}
  \country{China}
}

\author{Jinyan Feng}
\email{fjy@buaa.edu.cn}
\affiliation{%
  \institution{School of Transportation Science and Engineering, Beihang University‌}
  \city{Beijing}
  \country{China}
}

\author{Yuhui Liu}
\email{liuyh@csair.com}
\affiliation{%
  \institution{China Southern Airlines}
  \city{Guangzhou}
  \country{China}
}

\author{Zhongyi Pei}
\authornote{Corresponding Author.}
\email{peizhyi@tsinghua.edu.cn}
\affiliation{%
  \institution{School of Software, BNRist, \\ Tsinghua University}
  \city{Beijing}
  \country{China}
}
\author{Jianmin Wang}
\email{jimwang@tsinghua.edu.cn}
\affiliation{%
  \institution{School of Software, BNRist, \\ Tsinghua University}
  \city{Beijing}
  \country{China}
}

\renewcommand{\shortauthors}{Jiafeng Lin et al.}


\begin{abstract}
Predictive maintenance is vital for commercial aviation, yet progress remains hindered by the lack of public, large-scale real-world datasets. As a result, most methods are developed and evaluated on proprietary data, limiting reproducibility. We introduce \textbf{Pneuma}\footnote{Pneuma is an ancient Greek term for breath, air in motion, or vital spirit. The name reflects both the aircraft bleed air systems studied and the multivariate operational records that serve as vital signs of system health.}, the first publicly available large-scale benchmark for predictive maintenance using onboard time series from in-service commercial aircraft. Built around bleed air systems, Pneuma contains over 920 million multivariate time steps across four aircraft types and 9 verified fault scenarios, directly aligned with real maintenance logs. The dataset exhibits key real-world challenges, including substantial class imbalance, subtle fault evolution patterns, and significant variance across aircraft configurations. We structure the benchmark around three tasks: multivariate forecasting, supervised fault classification, and unsupervised anomaly detection, with labels derived from actual maintenance records. Under unified settings, we evaluate over 25 models from four methodological families on 9 carefully curated subsets. The results reveal that even strong contemporary models struggle with subtle fault signatures. Further analyses show consistent gains from additional in-context examples and multivariate inference, while forecasting quality is strongly associated with anomaly detection performance. Together, these findings provide practical insights into effectively modeling complex real-world aviation time series. We hope Pneuma will serve as a rigorous and reproducible foundation for advancing data-driven predictive maintenance toward real-world deployment in commercial aviation.\footnote{Preliminary work.}
\end{abstract}

\ccsdesc[500]{Information systems~Data mining}
\ccsdesc[500]{Mathematics of computing~Time series analysis}



\maketitle

\section{Introduction}

\begin{table*}[htbp]
    \caption{Comparison of representative publicly available aviation time series datasets and benchmarks. \emph{Commercial Operations} indicates that the data were collected from aircraft operating in commercial airline service. \emph{Standardized Benchmark} indicates that predefined tasks, data splits, and evaluation protocols are provided. \emph{Confirmed Fault Labels} indicates that the labels are associated with faults verified during actual aircraft operations or maintenance rather than simulated faults. \emph{Time Steps} denotes the total number of temporal observations included in the benchmark data.}
    \vspace{-7pt}
  \label{tab:aviation_comparison}
  \centering
  \begin{small}
    \begin{threeparttable}
      \setlength{\tabcolsep}{5pt}
      \resizebox{\textwidth}{!}{
        \begin{tabular}{l|cccccccc}
          \toprule
          Dataset & Commercial Aircraft & Benchmark & Modeling Task & Real-World Label & Aircraft Types & Variables & Fault Scenarios & Time Steps \\
          \midrule
          GATS \cite{labellagats} & {\transparent{0.5}\ding{55}} & \ding{51} & Classification, Reconstruction & {\transparent{0.5}\ding{55}} & 3 & 76 & 0 & $\sim$38.3M \\
          CAFUC2 \cite{jing2026high} & \transparent{0.5}\ding{55} & \ding{51} & Anomaly Detection & {\transparent{0.5}\ding{55}} & 4 & 22 & 4 & $\sim$1.0M \\
          NGAFID \cite{yang2022large} & {\transparent{0.5}\ding{55}} & \ding{51} & Classification & \ding{51} & 1 & 23 & 19 & $\sim$46.9M \\
          DASHlink \cite{dashlink_sample,dashlink_fourclass} & \ding{51} & \transparent{0.5}\ding{55} & Classification & \transparent{0.5}\ding{55} & 1 & 20 & 4 & $\sim$15.8M \\
          \midrule
          \textbf{Pneuma (Ours)} & \ding{51} & \ding{51} & Classification, Forecasting, Anomaly Detection & \ding{51} & \textbf{4} & \textbf{9$\sim$21} & \textbf{9} & \textbf{$\sim$921M} \\
          \bottomrule
        \end{tabular}
      }
    \end{threeparttable}
  \end{small}
  \vspace{-5pt}
\end{table*}

In commercial aviation, ensuring safety and operational reliability hinges on the timely detection of degradation in onboard systems. Unexpected system faults often lead to unscheduled maintenance, costly flight disruptions, and reduced aircraft availability, motivating data-driven predictive maintenance methods to identify early fault symptoms from operational telemetry. Modern aircraft continuously record multivariate sensor measurements throughout each flight, providing detailed observations of system behavior under routine operating conditions. However, these measurements are strongly affected by flight phases, operating environments, aircraft configurations, and fleet types, while early fault symptoms are often subtle and easily obscured by normal operational variation. Developing and evaluating reliable predictive maintenance models therefore requires representative onboard time series data collected from real-world airline operations, covering multiple aircraft types, diverse operating conditions, and practical fault scenarios.

Despite growing interest in aviation predictive maintenance, high-quality public benchmarks remain exceptionally scarce \cite{yang2022large}. Widely used aviation prognostics benchmarks primarily rely on simulated degradation processes or controlled experimental settings and therefore cannot fully reflect the variability of routine airline operations \cite{saxena2008damage}. While recent public datasets provide real-world flight logs \cite{labellagats,yang2022predictive}, they are largely restricted
to small general aviation fleets and therefore do not capture the complex system architectures or diverse fault modes of widely operated mainline aircraft such as the Airbus A320 and Boeing B777. Meanwhile, studies using operational data from commercial fleets are typically conducted on proprietary datasets with fragmented task definitions, arbitrary data splits, and inconsistent evaluation metrics \cite{su2023aircraft}. These limitations further hinder reproducibility, prevent fair cross-method comparison, and leave open the question of whether current time series models generalize across real-world aircraft configurations and fault scenarios. 

To bridge this gap, we introduce \textbf{Pneuma}, the first publicly available large-scale real-world dataset and benchmark for predictive maintenance derived from onboard time series from mainline commercial aircraft. Focusing on the aircraft Bleed Air System (BAS), a critical and failure-prone system regulating pneumatic pressure and temperature, \textbf{Pneuma} comprises over \emph{920 million time steps} from routine airline operations, spanning four aircraft types and 9 subsets defined by specific combinations of aircraft type and fault scenario. Crucially, all fault labels are derived from real maintenance logs and documented fault events, preserving the natural variation in flight phases, control states, and fault manifestations essential to realistic evaluation.

Around \textbf{Pneuma}, we organize three tasks that map directly to practical maintenance workflows and mainstream time 
series paradigms: \emph{multivariate forecasting} predicts future system states from earlier within-flight observations; \emph{supervised classification} identifies flight records with fault symptoms; and \emph{unsupervised anomaly detection} learns normal behavior and flags deviations linked to documented faults. 
Although these tasks differ in their supervisory and prediction objectives, they all evaluate model capabilities relevant to identifying fault
symptoms in aircraft operations. 
For each task, we provide standardized data splits, evaluation protocols, and metrics across all 9 subsets.
Under these unified settings, we systematically benchmark over 25 models from four methodological families: specialized forecasting, classification, anomaly detection, and few-shot time series foundation models, and conduct controlled experiments on representation and modeling choices across flight phases, configurations, and fault scenarios. Our comprehensive evaluation reveals that even state-of-the-art architectures struggle with subtle fault signatures, highlighting critical open challenges for the community.

Our contributions are summarized as follows:
\begin{itemize}[leftmargin=*, itemsep=0.2em]
    \item We introduce \textbf{Pneuma}, the first public large-scale real-world predictive maintenance dataset from in-service commercial aircraft, with over 920 million time steps, 4 aircraft types, and 9 real fault scenarios labeled from maintenance records.
    \item We establish a unified multi-task benchmark covering forecasting, classification, and anomaly detection, with standardized splits and evaluation protocols aligned to practical maintenance.
    \item Comprehensive evaluation of over 25 models reveals fundamental limitations of current methods and provides practical guidance on representation and modeling for aircraft time series.
\end{itemize}

\section{Related Work}

\subsection{Time Series Benchmarks}

Standardized benchmarks have been instrumental in driving the development and reliable evaluation of time series models \cite{wang2026deep, li2025tsfm}. For classification, the UCR and UEA archives \cite{dau2019ucr,bagnall2018uea} provide widely adopted collections of univariate and multivariate time series datasets with predefined evaluation splits. Benchmarking efforts for forecasting have gradually evolved from dataset aggregation to unified evaluation and, more recently, foundation model assessment. The M4 Competition \cite{makridakis2020m4} evaluates a broad range of statistical, machine learning, and hybrid forecasting methods on 100,000 real-world time series with diverse domains and sampling frequencies. TFB \cite{qiu2024tfb} further provides a unified and extensible pipeline for comparing statistical, machine learning, and deep learning forecasters under consistent experimental settings. More recently, TIME \cite{qiao2026s} introduces fresh datasets and operationally grounded forecasting tasks for strict zero-shot evaluation of time series foundation models. For anomaly detection, TSB-AD \cite{liu2024elephant} curates a large and heterogeneous collection of labeled time series and improves evaluation reliability through careful dataset inspection, unified experimental settings, and a systematic analysis of anomaly detection metrics.

While these general-purpose benchmarks facilitate broad comparisons across models, domain-specific benchmarks are needed to preserve the structures, labels, and evaluation requirements of individual applications. WeatherBench \cite{rasp2020weatherbench} establishes common data, metrics, and baselines for data-driven global weather forecasting. LargeST \cite{liu2023largest} captures the scale and long-term temporal coverage of real-world traffic networks while providing detailed sensor metadata. In predictive maintenance, BatteryLife \cite{tan2025batterylife} integrates battery degradation records across diverse chemistries, operating conditions, and experimental sources, enabling systematic comparison between specialized battery methods and general time series models. Together, these efforts demonstrate that a useful domain benchmark should combine representative data with clearly defined tasks, standardized evaluation protocols, and reproducible baselines. Following this principle, we establish a predictive maintenance benchmark tailored to the operational characteristics of onboard time series from commercial aircraft.

\subsection{Aviation Time Series Datasets}

The growing availability of flight-recorder data has led to a range of public aviation time series resources for prognostics, maintenance, and flight-safety research. C-MAPSS and N-CMAPSS \cite{saxena2008damage,arias2021aircraft} provide simulated turbofan-engine degradation trajectories and have become standard benchmarks for remaining useful life estimation. Among datasets collected from real flight operations, NASA DASHlink \cite{dashlink_sample,dashlink_fourclass} releases onboard recordings from a single type of regional jet operating in commercial service, including a curated final-approach dataset for multiclass anomaly detection. The NGAFID Aviation Maintenance Dataset \cite{yang2022predictive} associates general-aviation flight records with unplanned maintenance events, supporting maintenance classification and binary remaining useful life estimation. GATS \cite{labellagats} extends publicly available general-aviation data to three propeller-driven aircraft types and establishes benchmarks for aircraft classification and missing-data reconstruction. More recently, CAFUC2 \cite{jing2026high} combines recordings from four general-aviation aircraft models with physics-based fault injection to construct paired normal and anomalous samples for anomaly detection.
Table \ref{tab:aviation_comparison} compares the public real-world aviation time series datasets most relevant to our setting in terms of aviation domain, benchmark splits, label provenance, aircraft diversity, fault coverage, and dataset scale. Existing resources either focus on general aviation, cover a single regional-jet type, or obtain abnormal samples through synthetic fault injection. Consequently, a standardized public benchmark containing real-world, fault-associated onboard time series from multiple mainline commercial aircraft types remains lacking. Our dataset fills this gap with onboard time series from four mainline commercial aircraft types and unified benchmarks of aircraft bleed air systems.

\section{Pneuma}

\subsection{Dataset Overview}

\textbf{Pneuma} is a multivariate aircraft time series dataset constructed from Quick Access Recorder (QAR) measurements collected during routine airline operations, together with operational and maintenance records associated with documented fault events. Focused on critical Bleed Air Systems (BAS), \textbf{Pneuma} captures complex operational dynamics across four widely deployed mainline commercial aircraft: Airbus A320, A321, Boeing B787, and B777. Specifically, \textbf{Pneuma} contains 71,451 flight records and more than 920 million time steps collected between December 2021 and June 2026. It covers 9 aircraft-fault scenarios involving different system components, aircraft configurations, and fault manifestations. The dataset supports three benchmark tasks: multivariate forecasting, supervised classification, and unsupervised anomaly detection.

\begin{table*}[t]
    \centering
    \caption{
        Statistics of the nine aircraft-fault subsets.
        \emph{Collection Period} denotes the dates of the earliest and latest
        flight records in each subset.
        \emph{Variables} denotes the number of recorded QAR variables,
        excluding the time index and benchmark labels.
        \emph{Avg. Length} denotes the average number of time steps per
        flight trajectory.
        Flight records do not overlap across subsets.
    }
    \vspace{-7pt}
    \label{tab:dataset_statistics}
    \setlength{\tabcolsep}{4.2pt}
    \renewcommand{\arraystretch}{1.10}
    \small

    \resizebox{\textwidth}{!}{
    \begin{tabular}{lclcrrrr}
        \toprule
        \textbf{Subset}
        & \textbf{Aircraft Type}
        & \textbf{Target Scenario}
        & \textbf{Collection Period}
        & \textbf{Sensors}
        & \textbf{Flights}
        & \textbf{Avg. Length}
        & \textbf{Time Steps} \\
        \midrule

        320-DIFF  & A320 & 8HA sensing-line fault                         & 2024-01--2026-05 & 17 & 7,259 & 10.6K & 77.3M  \\
        320-HPV   & A320 & High-pressure valve fault                     & 2024-01--2026-05 & 17 & 8,681 & 10.7K & 92.7M  \\
        320-PRV   & A320 & Pressure-regulating valve fault               & 2024-01--2026-05 & 17 & 8,041 & 10.6K & 84.9M  \\
        320-WEAV  & A320 & Bleed-air duct leakage                       & 2024-01--2026-05 & 17 & 7,550 & 9.2K  & 69.2M  \\
        320-LOW   & A320 & PRV leakage with low transfer pressure        & 2021-12--2026-05 & 17 & 7,417 & 9.2K  & 68.2M  \\
        321-DIFF  & A321 & 8HA sensing-line fault                        & 2024-01--2026-05 & 17 & 8,894 & 11.1K & 98.8M  \\
        321-WEAV  & A321 & Bleed-air duct leakage                       & 2023-03--2026-05 & 17 & 8,052 & 9.0K  & 72.7M  \\
        787-CAC   & B787 & Cabin air compressor fault                    & 2023-01--2026-05 & 21 & 9,157 & 19.0K & 174.0M \\
        777-PRSOV & B777 & Pressure-regulating and shutoff valve fault   & 2024-01--2026-06 & 9  & 6,400 & 28.6K & 183.2M \\
        \midrule

        \textbf{Total}
        & \textbf{4 types}
        & \textbf{9 scenarios}
        & \textbf{2021-12--2026-06}
        & \textbf{9--21}
        & \textbf{71,451}
        & \textbf{12.9K}
        & \textbf{921.0M} \\

        \bottomrule
    \end{tabular}
    }
    \vspace{-5pt}
\end{table*}

\subsection{Operational Background}

Quick Access Recorders (QARs) continuously collect operational parameters from aircraft systems during flight \cite{wang2014analysis}. A QAR record contains synchronized measurements, such as pressures and temperatures, and control states, such as valve positions and flight phase information. Each flight therefore forms a multivariate time series recording how the aircraft and its onboard systems evolve from departure to arrival. Compared with isolated maintenance measurements, QAR data provide a continuous and detailed view of system behavior under actual operating conditions.

\begin{figure}[h]
\vspace{-5pt}
  \centering
  \includegraphics[width=0.95\linewidth]{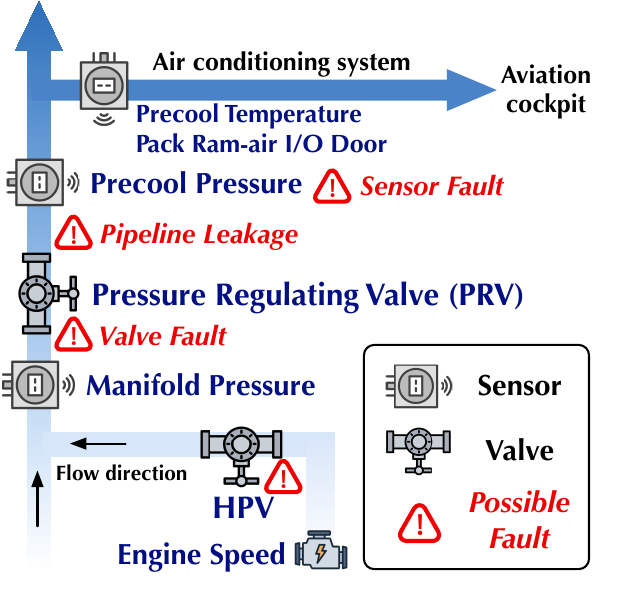}
  \vspace{-10pt}
  \caption{A320 Bleed Air System Schematic.}
  \label{fig:bleed_air_system}
  \vspace{-5pt}
\end{figure}

Figure \ref{fig:bleed_air_system} provides an example of aircraft bleed air system. In this system, compressed air from the engine is regulated through components such as the high pressure valve and pressure regulating valve before being supplied to downstream systems. Sensors continuously monitor the pressure, temperature, and operating state of this process. Faults may arise from valves, sensors, sensing lines, or air ducts, leading to abnormal pressure regulation, inconsistent sensor readings, or air leakage. While different aircraft may use different architectures, such as the conventional pneumatic system in the A320 and the electrically driven cabin compression system in the B787, they exhibit similar physical dependencies and generate comparable multivariate time series.


In routine operations, emerging faults rarely cause immediate or isolated spikes in a single sensor. Instead, symptoms manifest as subtle shifts, delayed responses, or cross-sensor inconsistencies that vary across flight phases and operating conditions. Furthermore, these symptoms can vary significantly depending on the current flight phase and environmental conditions. Predictive maintenance therefore seeks to identify these complex, early-stage patterns from QAR trajectories before they escalate into disruptive system faults. \textbf{Pneuma} pairs real-world operational trajectories with documented fault events, providing a practical benchmark for evaluating whether time series models can effectively capture these maintenance-relevant system behaviors.

\subsection{Dataset Collection}
\label{sec:dataset_collection}

The construction of \textbf{Pneuma} begins with establishing reliable ground truth. We first extract documented fault events from routine airline maintenance logs, identifying the affected aircraft, target component, event timestamp, and diagnostic description for each scenario. Based on these events, we retrieve the corresponding historical QAR telemetry and systematically filter out flights with missing variables or corrupted sensor streams. The remaining records are grouped by aircraft type and target fault scenario to form the 9 subsets summarized in Table~\ref{tab:dataset_statistics}. The collected QAR records are then processed based on two fundamental principles to make the data suitable for time series modeling.

\paragraph{Variable Selection} Modern aircraft continuously log thousands of parameters, most of which are irrelevant to the Bleed Air System. To remove background noise while preserving essential physical dynamics, we select a targeted subset of variables directly tied to system operation and fault mechanisms—specifically pressure, temperature, valve state, engine condition, and aircraft configuration.

\paragraph{Flight Phase Alignment} Commercial flights vary widely in duration from 1 to over 10 hours, making full-length raw sequences difficult to align or compare. Flight phase transitions involve control actions and frequent component actuation, producing informative responses in pressure, engine operating states, and valve commands.
Potential fault symptoms are therefore more likely to manifest through
abnormal system responses during these transitions. We therefore focus on four critical transitions: taxi-to-takeoff, takeoff-to-climb, climb-to-cruise, and cruise-to-descent. By extracting fixed-length time windows around these four transition points, we convert variable-length flights into standardized, comparable segments that highlight the most informative system responses.

\subsection{Benchmark}

\begin{figure*}[h]
  \centering
  \includegraphics[width=\linewidth]{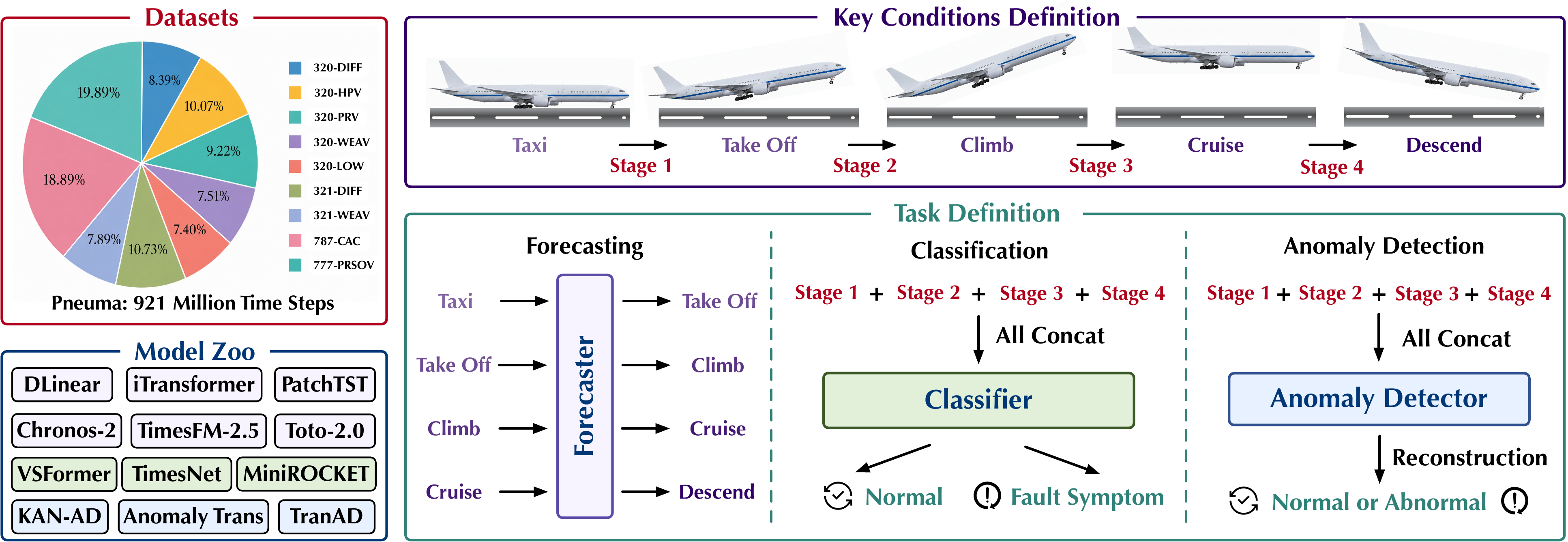}
  \vspace{-15pt}
  \caption{Overview of the Pneuma dataset and benchmark. The figure summarizes the dataset composition, key flight stages, task definitions for forecasting, classification, and anomaly detection, and the representative models evaluated in each track.}
  \label{fig:approach}
  \vspace{-5pt}
\end{figure*}

We construct the benchmark annotations relative to documented fault events, as shown in Figure~\ref{fig:approach}. For each aircraft and target system, flights are ordered chronologically, and each confirmed fault event is used as a temporal reference point. We define two sample pools representing normal operation and the period in which fault symptoms may emerge.

The \emph{fault warning pool} comprises the 100 consecutive flights immediately preceding each fault event. In operational observations, fault-related symptoms tend to grow more frequent and pronounced as the failure approaches; the 100‑flight window captures this period of intensifying precursor signals. At the same time, a lead time of 100 flights is regarded as operationally meaningful, providing sufficient advance notice for inspection, maintenance planning, and decision-making. The window therefore balances symptom relevance with practical warning utility.
The \emph{normal pool} consists of flights that are sufficiently separated from all documented fault events. To reduce contamination from early degradation or post‑maintenance transients, we exclude a margin of 500 flights before and 50 flights after each event. The resulting gap between the warning window (100 flights) and the exclusion boundary (500 flights) forms a 400‑flight temporal buffer, which enforces a cleaner separation between normal and pre‑fault conditions. These annotation rules were established and validated through operational maintenance practice. The two pools are assigned distinct roles across the benchmark tasks, as detailed below.

\subsubsection{Forecasting}

We formulate flight phase transition forecasting as a complementary
paradigm for predictive maintenance. Phase transitions involve coordinated
changes in component actuation, pressure, temperature, airflow, and control
states, providing informative operating conditions for observing system
dynamics. Accurate prediction across these transitions therefore reflects
a model's ability to capture the normal responses and cross-variable
relationships of aircraft systems.

Each flight yields four forecasting cases corresponding to the transitions
defined in Section~\ref{sec:dataset_collection}. For each case, the
multivariate observations are used to predict the
system states after the transition. Models are trained, validated, and
tested exclusively on the normal pool to focus on healthy operational
dynamics.

Although forecasting does not directly predict fault labels, fault
development can alter normal system responses during phase transitions.
Thus, representations learned through forecasting may also support fault
recognition. This connection is empirically supported in
Section \ref{sec:forecasting_classification_relation}, where models with
better forecasting performance generally achieve stronger anomaly detection
results under the same normal-only reconstruction setting.
We evaluate forecasting using Mean Absolute Error (MAE) and Root Mean
Squared Error (RMSE), with lower values indicating better performance.

\subsubsection{Classification}

The classification task evaluates whether a model can identify fault symptoms from complete flight records. The four aligned transition segments of each flight are concatenated in chronological order to form a flight-level multivariate time series. Flights in the normal pool are labeled as normal, whereas flights in the fault warning pool are labeled as warning samples. Both classes are included in the training, validation, and test sets.

In operational aviation settings, normal flights far outnumber those exhibiting fault symptoms. Excessive false alarms can impose prohibitive inspection and maintenance costs, even when some impending faults are detected; a practical warning system must therefore balance detection coverage against alarm precision. Based on maintenance practice, we set a minimum recall target of $10\%$ for the fault warning class: detecting 10 of the 100 flights leading up to a fault provides actionable lead time for preventive maintenance. We sweep the decision threshold and select the first operating point at which warning recall reaches or exceeds $10\%$. The corresponding precision, denoted P@R10, serves as our evaluation metric. It directly measures warning reliability at a predefined operational detection coverage, with higher values indicating fewer false alarms.

\subsubsection{Anomaly Detection}

The anomaly detection task evaluates whether fault symptoms can be flagged without using labeled warning samples during training. Utilizing the same flight-level multivariate time series representation, models are trained exclusively on normal pool flights to learn reference operational behavior. The test set contains both normal and warning flights. Following the reconstruction-based anomaly detection setting, each model reconstructs the input series, and the averaged reconstruction error across time steps and variables is used as the flight-level anomaly score. Higher scores indicate greater deviation from the normal behavior observed during training. To maintain a consistent operational interpretation across the two fault detection tasks, anomaly detection methods are evaluated using the same $P@R10$ metric as the supervised classification methods.

\section{Experiments}

\subsection{Experimental Setup}

\newcommand{\smallhead}[1]{{\footnotesize #1}}

\begin{table*}[t]
    \centering
    \caption{
    Main forecasting results on nine aircraft system datasets. The first group reports full-shot models trained separately on each dataset, while the second group reports time series foundation models evaluated in the 5-shot setting. Each entry reports $\mathrm{Error}=(\mathrm{MAE}+\mathrm{RMSE})/2$. Lower Error and higher Score indicate better performance. The best result is highlighted in \textbf{bold}.
    }
    \vspace{-7pt}
    \label{tab:main_forecasting}
    \setlength{\tabcolsep}{2.2pt}
    \renewcommand{\arraystretch}{1.12}
    \small

    \resizebox{\textwidth}{!}{
    \begin{tabular}{l|*{9}{c}|c}
        \toprule
        \textbf{Model}
        & \smallhead{\textbf{320-DIFF}}
        & \smallhead{\textbf{320-HPV}}
        & \smallhead{\textbf{320-PRV}}
        & \smallhead{\textbf{320-WEAV}}
        & \smallhead{\textbf{320-LOW}}
        & \smallhead{\textbf{321-DIFF}}
        & \smallhead{\textbf{321-WEAV}}
        & \smallhead{\textbf{787-CAC}}
        & \smallhead{\textbf{777-PRSOV}}
        & \textbf{Score} \\
        \midrule

        \rowcolor{gray!15}
        \multicolumn{11}{c}{
            \textit{\textbf{Time Series Forecasting Models (Full-shot)}}
        } \\
        \midrule

        OLinear \cite{yue2026olinear}
        & 0.411 & 0.416 & 0.414 & 0.448 & 0.446
        & 0.384 & 0.460 & \textbf{0.476} & 0.458 & 69.72 \\

        xPatch \cite{stitsyuk2025xpatch}
        & 0.410 & 0.415 & 0.414 & 0.442 & 0.439
        & 0.381 & 0.459 & 0.492 & 0.451 & 69.78 \\

        DUET \cite{qiu2025duet}
        & 0.401 & 0.404 & 0.403 & 0.438 & 0.435
        & 0.371 & 0.455 & 0.475 & 0.461 & 70.11 \\

        TimeMixer \cite{wang2024timemixer}
        & 0.451 & 0.456 & 0.453 & 0.498 & 0.497
        & 0.428 & 0.484 & 0.542 & 0.490 & 67.70 \\

        TimeXer \cite{wang2024timexer}
        & 0.428 & 0.432 & 0.430 & 0.475 & 0.471
        & 0.401 & 0.474 & 0.487 & 0.478 & 68.86 \\

        iTransformer \cite{liu2024itransformer}
        & 0.419 & 0.424 & 0.422 & 0.452 & 0.449
        & 0.391 & 0.473 & 0.505 & 0.460 & 69.29 \\

        DLinear \cite{zeng2023transformers}
        & \textbf{0.266} & \textbf{0.266} & \textbf{0.267} & \textbf{0.304} & \textbf{0.302}
        & \textbf{0.243} & \textbf{0.274} & 0.537 & \textbf{0.325} & \textbf{76.65} \\

        PatchTST \cite{nie2022time}
        & 0.425 & 0.433 & 0.430 & 0.457 & 0.455
        & 0.400 & 0.481 & 0.523 & 0.466 & 68.90 \\

        TimesNet \cite{wu2022timesnet}
        & 0.404 & 0.407 & 0.406 & 0.453 & 0.449
        & 0.379 & 0.465 & 0.457 & 0.455 & 69.93 \\

        Autoformer \cite{wu2021autoformer}
        & 0.519 & 0.513 & 0.521 & 0.583 & 0.584
        & 0.480 & 0.553 & 0.754 & 0.564 & 64.10 \\

        \midrule
        \rowcolor{gray!15}
        \multicolumn{11}{c}{
            \textit{\textbf{Time Series Foundation Models (5-shot)}}
        } \\
        \midrule

        TiRex-2 \cite{podest2026tirex}
        & 0.401 & 0.409 & 0.403 & 0.423 & 0.421
        & 0.384 & 0.454 & 0.490 & 0.459 & 70.12 \\
        
        Chronos-2 \cite{ansari2025chronos}
        & 0.399 & 0.407 & 0.402 & 0.407 & 0.405
        & 0.383 & 0.441 & \textbf{0.447} & 0.447 & 70.67 \\
        TimesFM-2.5 \cite{das2023decoder}
        & 0.414 & 0.421 & 0.417 & 0.451 & 0.448
        & 0.400 & 0.465 & 0.518 & 0.476 & 69.22 \\
        Toto-2.0 \cite{khwaja2026toto}
        & \textbf{0.389} & \textbf{0.397} & \textbf{0.392} & \textbf{0.404} & \textbf{0.401}
        & \textbf{0.373} & \textbf{0.433} & 0.463 & \textbf{0.438} & \textbf{70.95} \\
        Moirai-2.0 \cite{liu2025moirai}
        & 0.447 & 0.454 & 0.449 & 0.489 & 0.488
        & 0.440 & 0.518 & 0.546 & 0.520 & 67.45 \\
        \bottomrule
    \end{tabular}
    }
    \vspace{-5pt}
\end{table*}

\paragraph{Baselines.}

To comprehensively evaluate \textbf{Pneuma}, we benchmark over 25 representative models across the three task paradigms.

For the forecasting task, we first evaluate 10 well-acknowledged time series forecasting models trained separately on each dataset, including
Autoformer~\cite{wu2021autoformer}, TimesNet~\cite{wu2022timesnet}, DLinear~\cite{zeng2023transformers}, PatchTST~\cite{nie2022time}, iTransformer~\cite{liu2024itransformer}, TimeXer~\cite{wang2024timexer}, TimeMixer~\cite{wang2024timemixer}, DUET~\cite{qiu2025duet}, and recent xPatch~\cite{stitsyuk2025xpatch} and OLinear~\cite{yue2026olinear}.
We also evaluate recent time series foundation models strictly in a 5-shot setting, without fine-tuning on the specific dataset. The evaluated models include TimesFM-2.5~\cite{das2023decoder}, Moirai-2.0~\cite{liu2025moirai}, Chronos-2~\cite{ansari2025chronos}, TiRex-2~\cite{podest2026tirex} and Toto-2.0~\cite{khwaja2026toto}.

For classification, we include specialized classifiers MambaSL~\cite{jung2026mambasl}, VSFormer~\cite{xi2024vsformer}, LITE~\cite{ismail2025look}, together with the convolutional-kernel methods MiniROCKET~\cite{dempster2021minirocket} and MultiROCKET~\cite{tan2022multirocket}. We also adapt the general time series architectures TimesNet~\cite{wu2022timesnet}, PatchTST~\cite{nie2022time}, DLinear~\cite{zeng2023transformers}, and iTransformer~\cite{liu2024itransformer} to the classification setting by replacing their original prediction heads with classification heads.

For anomaly detection, we evaluate KAN-AD~\cite{zhou2024kan}, Anomaly Transformer~\cite{xu2021anomaly}, TranAD~\cite{tuli2022tranad}, USAD~\cite{audibert2020usad}, and OmniAnomaly~\cite{su2019robust}. These methods cover several major paradigms for multivariate time series anomaly detection, including reconstruction-based detection, adversarial autoencoding, probabilistic latent-variable modeling, attention-based association analysis, and prediction or reconstruction with Transformer architectures.

\paragraph{Implementation Details.}

All experiments are implemented in PyTorch \cite{paszke2019pytorch} and conducted on NVIDIA GeForce RTX 4090 GPUs. To ensure a fair comparison, all trainable models follow the same data pre-processing procedure and use unified training and optimization settings unless otherwise specified. Detailed sample statistics, data
partitioning protocols, and task-specific input and output lengths are
provided in Appendix \ref{sec:dataset_details}. Definitions for all evaluation metrics  are provided in Appendix \ref{sec:implementation_details}.

\vspace{-5pt}
\subsection{Main Results}

Table \ref{tab:main_forecasting} summarizes the forecasting performance on the nine datasets, with the complete MAE and RMSE results provided in Appendix \ref{sec:full_forecasting_results} and the computation of \emph{Score} detailed in Appendix \ref{sec:implementation_details}. Among the full-shot forecasting models, DLinear achieves the highest overall Score of 76.65 and obtains the lowest forecasting error on eight of the nine datasets. Its strong performance compared with more complex Transformer and multiscale architectures suggests that simple temporal mappings remain highly competitive for the operational patterns represented in Pneuma. TimesNet performs best on 787-CAC among the full-shot baselines, indicating that the relative effectiveness of different architectures can vary across aircraft and system configurations.

The time series foundation models also show competitive performance under the 5-shot setting. Toto-2.0 achieves the best overall performance among the foundation models and ranks first on eight datasets, while Chronos-2 obtains the best result on the remaining dataset. Although the strongest foundation models outperform most full-shot baselines, none surpasses DLinear in terms of the aggregate Score. Nevertheless, their strong performance without dataset-specific training demonstrates an ability to exploit pretrained temporal knowledge from only a small number of contextual examples. Forecasting errors remain substantially higher on 787-CAC than on most other datasets, suggesting that its operational dynamics pose a greater modeling challenge. Overall, these results highlight the potential of pretrained foundation models for aviation time series forecasting, while also showing that specialized models trained on target data remain highly competitive.

\begin{table*}[t]
    \centering
    \caption{
    Main results for supervised fault classification and unsupervised anomaly detection on nine aircraft fault datasets. Classification models are trained with labeled normal and fault-symptom samples, whereas anomaly detection models learn only from normal samples. All entries report Precision@R10, and \emph{Score} denotes the aggregate performance across datasets. Higher values indicate better performance, and the best result in each column within each track is highlighted in bold.
    }
    \vspace{-7pt}
    \label{tab:main_results_classification_anomaly}
    \setlength{\tabcolsep}{2.2pt}
    \renewcommand{\arraystretch}{1.12}
    \small

    \resizebox{\textwidth}{!}{
    \begin{tabular}{l|*{9}{c}|c}
        \toprule
        \textbf{Model}
        & \smallhead{\textbf{320-DIFF}}
        & \smallhead{\textbf{320-HPV}}
        & \smallhead{\textbf{320-PRV}}
        & \smallhead{\textbf{320-WEAV}}
        & \smallhead{\textbf{320-LOW}}
        & \smallhead{\textbf{321-DIFF}}
        & \smallhead{\textbf{321-WEAV}}
        & \smallhead{\textbf{787-CAC}}
        & \smallhead{\textbf{777-PRSOV}}
        & \textbf{Score} \\
        \midrule

        \rowcolor{gray!15}
        \multicolumn{11}{c}{
            \textit{\textbf{Classification Models}}
        } \\
        \midrule

        MambaSL \cite{jung2026mambasl}
        & \textbf{1.000} & 0.303 & 0.275 & 0.576 & 0.163
        & 0.113 & 0.233 & 0.287 & 0.444
        & 37.71 \\

        VSFormer \cite{xi2024vsformer}
        & \textbf{1.000} & \textbf{0.649} & \textbf{0.364} & \textbf{0.655} & 0.366
        & 0.564 &\textbf{ 0.593} & 0.288 & 0.941
        & \textbf{60.22} \\

        LITE \cite{ismail2025look}
        & \textbf{1.000} & 0.143 & 0.128 & 0.333 & \textbf{0.577}
        & \textbf{1.000} & 0.270 & 0.330 & \textbf{1.000}
        & 53.12 \\

        TimesNet \cite{wu2022timesnet}
        & 0.955 & 0.325 & 0.220 & 0.413 & 0.333
        & \textbf{1.000} & 0.174 & 0.323 & 0.182
        & 43.61 \\

        PatchTST \cite{nie2022time}
        & 0.949 & 0.202 & 0.220 & 0.625 & 0.283
        & 0.231 & 0.234 & \textbf{0.587} & 0.435
        & 41.84 \\

        DLinear \cite{zeng2023transformers}
        & 0.778 & 0.545 & 0.200 & 0.373 & 0.228
        & 0.611 & 0.190 & 0.182 & 0.205
        & 36.80 \\

        iTransformer \cite{liu2024itransformer}
        & 0.778 & 0.118 & 0.245 & 0.455 & 0.231
        & 0.367 & 0.217 & 0.244 & 0.262
        & 32.41 \\

        MultiROCKET \cite{tan2022multirocket}
        & 0.913 & 0.313 & 0.203 & 0.352 & 0.283
        & \textbf{1.000} & 0.223 & 0.156 & 0.571
        & 44.60 \\

        MiniROCKET \cite{dempster2021minirocket}
        & 0.913 & 0.237 & 0.232 & 0.322 & 0.319
        & \textbf{1.000} & 0.226 & 0.190 & 0.258
        & 41.08 \\

        \midrule

        \rowcolor{gray!15}
        \multicolumn{11}{c}{
            \textit{\textbf{Anomaly Detection Models}}
        } \\
        \midrule

        KAN-AD \cite{zhou2024kan}
        & 0.137 & 0.129 & \textbf{0.305} & \textbf{0.292} & 0.263
        & 0.077 & 0.330 & 0.230 & 0.165
        & 21.42 \\

        Anomaly Trans. \cite{xu2021anomaly}
        & \textbf{0.184} & \textbf{0.431} & 0.250 & 0.238 & 0.185
        & \textbf{0.101} & \textbf{0.341} & 0.284 & 0.066
        & \textbf{23.11} \\

        TranAD \cite{tuli2022tranad}
        & 0.093 & 0.093 & 0.288 & 0.132 & 0.095
        & 0.044 & 0.286 & 0.302 & \textbf{0.471}
        & 20.04 \\

        USAD \cite{audibert2020usad}
        & 0.102 & 0.125 & 0.224 & 0.275 & 0.333
        & 0.069 & 0.222 & \textbf{0.376} & 0.174
        & 21.11 \\

        OmniAnomaly \cite{su2019robust}
        & 0.059 & 0.129 & 0.151 & 0.235 & \textbf{0.341}
        & 0.070 & 0.204 & 0.311 & 0.271
        & 19.68 \\

        \bottomrule
    \end{tabular}
    }
    \vspace{-5pt}
\end{table*}

Table \ref{tab:main_results_classification_anomaly} reports the classification and anomaly detection results on the nine datasets. Among the supervised classification methods, VSFormer achieves the highest overall Score of 60.22 and performs best on several datasets. LITE ranks second overall and performs particularly well on 320-LOW and 777-PRSOV, while PatchTST achieves the strongest result on 787-CAC. These results indicate that no single architecture consistently dominates across all aircraft and fault scenarios. Specialized classification models achieve the strongest aggregate performance, while forecasting architectures adapted for classification remain competitive on individual datasets.

The anomaly detection methods perform substantially worse than the supervised classifiers, with the best model, Anomaly Transformer, achieving an overall Score of only 23.11. In our reconstruction-based setting, models are trained exclusively on normal samples and detect anomalies using reconstruction errors at test time, following the common assumption that abnormal patterns are more difficult to reconstruct. However, this assumption is challenging for aircraft operational data. Normal records exhibit substantial variation across flight phases and operating conditions, while early fault symptoms are often weak and partially overlap with normal behavior. Consequently, normal and fault-symptom samples may produce similar reconstruction errors, limiting their separability. These results suggest that learning normality alone is insufficient for complex aviation faults and that limited fault supervision may be necessary for more reliable detection.

\subsection{Analysis}

\subsubsection{Effect of In-Context Learning}

We investigate how the number of in-context examples affects the forecasting performance of time series foundation models. For each target flight, the preceding $n$ complete flight trajectories are provided as n-shot demonstrations, allowing the model to observe the full system evolution across the corresponding flight phase transition. We vary the number of demonstrations among $2$, $3$, $5$, $8$, and $12$, while keeping all other evaluation settings unchanged.

\begin{figure}[t]
  \centering
  \includegraphics[width=0.95\linewidth]{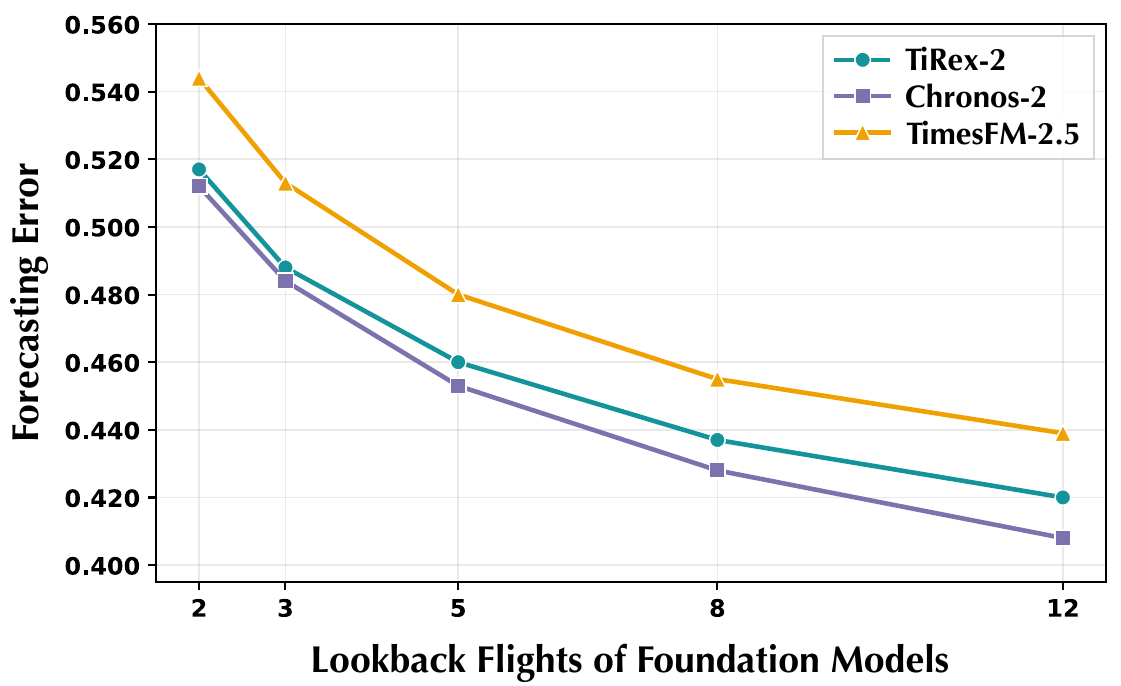}
  \vspace{-8pt}
  \caption{Effect of the number of in-context flight demonstrations. Each demonstration contains the complete multivariate trajectory of a preceding flight.}
  \label{fig:lookback}
  \vspace{-15pt}
\end{figure}

As shown in Figure~\ref{fig:lookback}, the forecasting loss decreases consistently with the number of demonstrations for all three foundation models. This monotonic improvement indicates that the models can extract useful information from multiple complete flight examples and adapt their predictions to the recent operating characteristics of the aircraft system. These results also show that historical flights in \textbf{Pneuma} contain consistent and reusable operational patterns. Complete trajectories from preceding flights provide useful references for predicting the target flight, which supports the coherence of the data collection and processing pipeline.

\subsubsection{Univariate versus Multivariate Inference}

\begin{figure}[h]
  \centering
  \includegraphics[width=\linewidth]{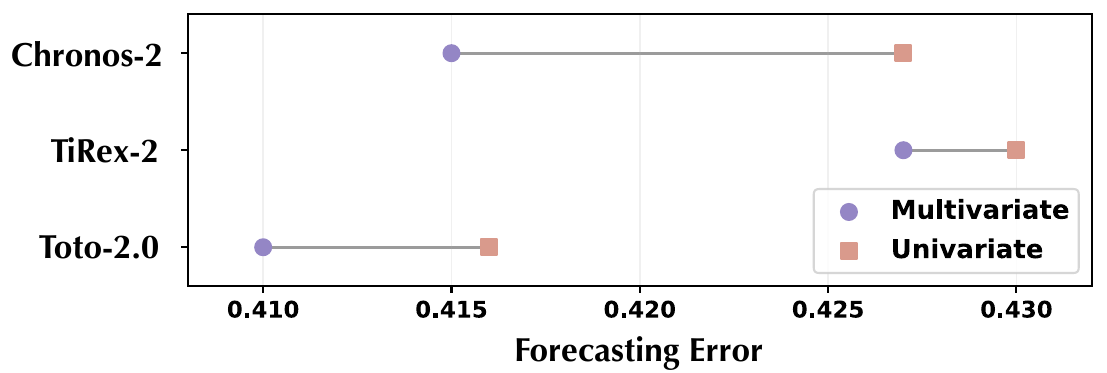}
  \vspace{-15pt}
  \caption{
    Comparison of univariate and multivariate inference for three time series foundation models.}
  \label{fig:multi_vs_uni}
  \vspace{-15pt}
\end{figure}

We further investigate whether jointly modeling multiple variables benefits pretrained time series foundation models. We select Chronos-2, TiRex-2, and Toto-2.0, all of which support multivariate inference, and compare two settings. In the univariate setting, each variable is forecast independently, requiring separate inference for all variables in a sample. In the multivariate setting, all variables are provided jointly and predicted in a single inference pass. As shown in Figure \ref{fig:multi_vs_uni}, multivariate inference consistently reduces forecasting loss for all three models. The consistent improvement provides empirical evidence that the
variables in Pneuma contain informative cross-variable dependencies, which is consistent with the physical coupling among components and control signals in aircraft systems. It also demonstrates the value of multivariate pretraining, as jointly trained foundation models can exploit interactions that are unavailable when each variable is processed independently.

\subsubsection{Effect of Temporal Patch Length}
\label{sec:patch_length}

\begin{figure}[h]
  \centering
  \includegraphics[width=0.95\linewidth]{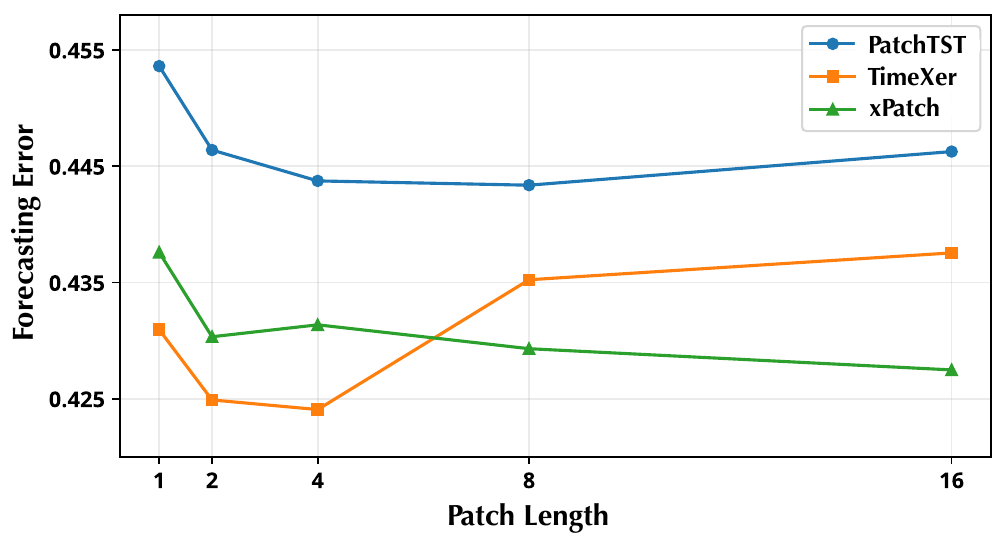}
  \vspace{-5pt}
  \caption{Effect of patch length on forecasting performance.}
  \label{fig:patch_length}
  \vspace{-5pt}
\end{figure}

Patch-based forecasting models represent consecutive observations as temporal
tokens, making patch length an important factor in determining the temporal
granularity of the learned representation. To investigate how aircraft
operational time series should be segmented, we vary the patch length among
$1$, $2$, $4$, $8$, and $16$ for PatchTST, TimeXer, and xPatch, while keeping
all other experimental settings unchanged. Figure \ref{fig:patch_length}
reports the forecasting loss, computed as the average of MAE and RMSE and
further averaged across the evaluated subsets and four flight phase
transitions. The complete numerical results are reported in Appendix \ref{app:patch_length_results}.

All three models benefit from increasing the patch length beyond the
point-level setting of $1$, showing that grouping adjacent observations
provides a more effective representation of aircraft system dynamics.
However, the optimal temporal granularity varies across architectures.
For PatchTST and TimeXer, performance deteriorates when
the patches become longer than their respective optimal settings, suggesting
that excessive temporal aggregation may obscure short duration responses
around flight phase transitions. Conversely, very short patches may fragment
coherent component responses and provide insufficient context for modeling
system evolution. Overall, the results indicate that aircraft time series
benefit from patch based temporal aggregation, while the appropriate patch
length depends on how each model represents and processes temporal patterns.

\subsubsection{Relationship between Forecasting and Anomaly Detection.}
\label{sec:forecasting_classification_relation}

\begin{figure}[t]
  \centering
  \includegraphics[width=\linewidth]{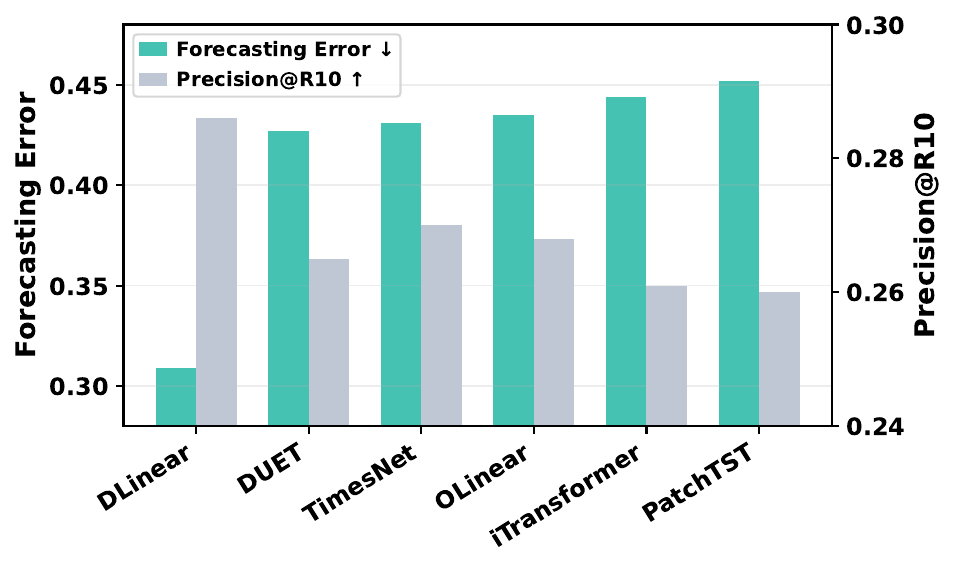}
  \vspace{-23pt}
  \caption{Relationship between forecasting and anomaly detection performance across six representative architectures.}
  \label{fig:vs}
  \vspace{-10pt}
\end{figure}

We examine whether forecasting capability is associated with anomaly detection performance. Specifically, we train each model under the same reconstruction-based anomaly detection setting using only normal samples, and use the averaged reconstruction error as the anomaly score. Figure \ref{fig:vs} compares the average forecasting error and Precision@R10 of six representative architectures. A clear performance consistency can be observed: DLinear achieves both the lowest forecasting error of 0.309 and the highest Precision@R10 of 0.286, while anomaly detection performance generally decreases as forecasting error increases (Pearson's $r=-0.954$, $p=0.003$). Since lower forecasting error and higher Precision@R10 both reflect better performance, this negative correlation indicates a positive association between forecasting and anomaly detection capabilities. This finding suggests that models capable of accurately capturing system dynamics are also more likely to distinguish subtle deviations from normal operational behavior.

\section{Conclusion and Future Work}

In this work, we introduce \textbf{Pneuma}, a large-scale public benchmark for predictive maintenance using onboard time series collected from in-service mainline commercial aircraft. Pneuma contains more than 920 million time steps across four aircraft types and 9 fault scenarios derived from the aircraft bleed air system. We formulate three time series tasks, including multivariate forecasting, supervised classification, and unsupervised anomaly detection, and establish standardized data construction, splitting, and evaluation protocols. Our comprehensive benchmarking of over 25 models reveals that existing time series models still face substantial difficulties in recognizing early fault symptoms and modeling diverse aircraft
configurations.

Future work will extend Pneuma in both data coverage and modeling scope. We plan to include more aircraft types, operational records, onboard systems, and fault scenarios to improve the diversity and representativeness of the benchmark. On the modeling side, promising directions include learning from limited fault samples, transferring knowledge across aircraft types and systems, and developing more general time series models for aviation maintenance. We hope Pneuma will support continued research toward reliable and transferable predictive maintenance methods.

\bibliographystyle{ACM-Reference-Format}
\bibliography{sample-base}

\appendix

\section{Details of Datasets}
\label{sec:dataset_details}

\paragraph{Data Organization.}
Each subset consists of multivariate QAR trajectories collected at the
flight level and is organized into normal and abnormal samples according
to the corresponding fault scenario. A complete flight record is treated
as the basic unit for sample construction and data partitioning. For each documented fault event, the associated flight records are chronologically divided into training, validation, and test sets with a ratio of 7:2:1. We refer to this protocol as a within-event chronological split, which preserves the temporal order and prevents
future flights from being included in earlier splits.

\paragraph{Data Format.}
Pneuma is distributed in the Apache TsFile format
\cite{zhao2024apache}, an open-source columnar storage format designed
for large-scale time series data. TsFile supports efficient compression,
high-throughput reading and writing, and indexed access by variable and
time range. These properties reduce the storage and data loading overhead
of large multivariate QAR records and facilitate efficient sample
extraction for downstream learning tasks.

\paragraph{Task-specific Sample Construction.}
For forecasting, we extract a fixed-length sequence around each selected
flight phase transition. The preceding 60 time steps are used as input,
and the subsequent 20 time steps are used as the prediction target. For
classification and anomaly detection, the segments selected from the four
flight stage transitions are concatenated to form a flight-level sequence. The stage
segments are determined using an aviation-specific operating-condition
selection protocol, and their lengths therefore differ across aircraft
types.

\paragraph{Sample Distribution.}
Table~\ref{tab:sample_task_statistics} reports the numbers of normal and
abnormal flights together with the task-specific sequence lengths. The
benchmark contains 71,451 flights in total, including 58,141 normal
samples and 13,310 abnormal samples. Normal flights substantially
outnumber abnormal flights across all subsets, reflecting the practical
distribution of airline operations, in which fault-related records are
naturally scarce. We retain this class imbalance to provide a realistic
evaluation of predictive maintenance methods.

\begin{table}[t]
    \centering
    \caption{
        Flight-level sample distribution and task-specific sequence lengths
        of the nine aircraft fault subsets. Forecasting uses 60 input time
        steps to predict the subsequent 20 time steps. The final column
        reports the concatenated sequence length used for classification
        and anomaly detection.
    }
    \label{tab:sample_task_statistics}
    \setlength{\tabcolsep}{3.2pt}
    \renewcommand{\arraystretch}{1.10}
    \small

    \resizebox{\columnwidth}{!}{
    \begin{tabular}{lrrrrr}
        \toprule
        \textbf{Subset}
        & \textbf{Normal}
        & \textbf{Abnormal}
        & \textbf{Input}
        & \textbf{Output}
        & \textbf{Cls./AD Length} \\
        \midrule

        320-DIFF  & 6,186 & 1,073 & 60 & 20 & 2,235 \\
        320-HPV   & 6,186 & 2,495 & 60 & 20 & 2,235 \\
        320-PRV   & 6,186 & 1,855 & 60 & 20 & 2,235 \\
        320-WEAV  & 6,300 & 1,250 & 60 & 20 & 2,235 \\
        320-LOW   & 6,300 & 1,117 & 60 & 20 & 2,235 \\
        321-DIFF  & 7,797 & 1,097 & 60 & 20 & 2,235 \\
        321-WEAV  & 6,700 & 1,352 & 60 & 20 & 2,235 \\
        787-CAC   & 6,886 & 2,271 & 60 & 20 & 2,920 \\
        777-PRSOV & 5,600 &   800 & 60 & 20 & 2,235 \\
        \midrule

        \textbf{Total}
        & \textbf{58,141}
        & \textbf{13,310}
        & \textbf{60}
        & \textbf{20}
        & \textbf{2,235 / 2,920} \\

        \bottomrule
    \end{tabular}
    }
\end{table}

\section{Definition of Aviation Variables}

The recorded variables in Pneuma describe the operating conditions and
component states of the aircraft bleed air system. Since variable names
and logging conventions differ across aircraft types, we present the
variables of the A320 subsets as a representative example. The five A320
subsets, including 320-DIFF, 320-HPV, 320-PRV, 320-WEAV, and 320-LOW,
share the same set of 17 QAR variables. The time index and benchmark
labels are not included in this list.

The suffixes ``1'' and ``2'' refer to the systems associated with
Engine~1 and Engine~2, respectively. Variables ending in ``\_R'' record
valve status feedback, while ``I'' and ``O'' denote the inlet and outlet
sides of the corresponding component. Table \ref{tab:a320_variables}
summarizes the physical meaning of each variable.

\begin{table*}[t]
    \centering
    \caption{
        Definitions of the QAR variables used in the A320 subsets.
        The time index and benchmark labels are excluded.
    }
    \label{tab:a320_variables}
    \setlength{\tabcolsep}{8pt}
    \renewcommand{\arraystretch}{1.12}
    \small

    \begin{tabular}{p{0.24\textwidth} p{0.69\textwidth}}
        \toprule
        \textbf{Variable} & \textbf{Physical Meaning} \\
        \midrule

        \texttt{N21}
        & High-pressure rotor speed of Engine~1, representing the operating
        state of the engine bleed air source. \\

        \texttt{N22}
        & High-pressure rotor speed of Engine~2. \\

        \texttt{BMPS1}
        & Bleed manifold pressure associated with Engine~1. \\

        \texttt{BMPS2}
        & Bleed manifold pressure associated with Engine~2. \\

        \texttt{PRECOOL\_PRESS1}
        & Bleed air pressure measured at the precooler section associated
        with Engine~1, downstream of the pressure regulating valve. \\

        \texttt{PRECOOL\_PRESS2}
        & Bleed air pressure measured at the precooler section associated
        with Engine~2. \\

        \texttt{FLIGHT\_PHASE}
        & Categorical indicator of the current flight phase. \\

        \texttt{PRV\_ENG1\_R}
        & Status feedback of the pressure regulating valve associated with
        Engine~1. \\

        \texttt{PRV\_ENG2\_R}
        & Status feedback of the pressure regulating valve associated with
        Engine~2. \\

        \texttt{HPV\_ENG1\_R}
        & Status feedback of the high-pressure bleed valve associated with
        Engine~1. \\

        \texttt{HPV\_ENG2\_R}
        & Status feedback of the high-pressure bleed valve associated with
        Engine~2. \\

        \texttt{PRECOOL\_TEMP1}
        & Bleed air temperature at the precooler outlet associated with
        Engine~1. \\

        \texttt{PRECOOL\_TEMP2}
        & Bleed air temperature at the precooler outlet associated with
        Engine~2. \\

        \texttt{PACK1\_RAM\_I\_DR}
        & Position of the ram-air inlet door for air-conditioning Pack~1. \\

        \texttt{PACK1\_RAM\_O\_DR}
        & Position of the ram-air outlet door for air-conditioning Pack~1. \\

        \texttt{PACK2\_RAM\_I\_DR}
        & Position of the ram-air inlet door for air-conditioning Pack~2. \\

        \texttt{PACK2\_RAM\_O\_DR}
        & Position of the ram-air outlet door for air-conditioning Pack~2. \\

        \bottomrule
    \end{tabular}
\end{table*}

\section{Implementation Details}
\label{sec:implementation_details}

All experiments are implemented in PyTorch
\cite{paszke2019pytorch} and conducted on NVIDIA GeForce RTX 4090 GPUs.
To ensure fair comparison, all trainable models within the same track
follow identical data preprocessing and sample construction procedures
and use unified training and optimization settings unless otherwise
specified. Detailed sample statistics, data partitioning protocols, and
task-specific input and output lengths are provided in
Appendix~\ref{sec:dataset_details}.

We evaluate forecasting performance using mean absolute error (MAE) and
root mean squared error (RMSE). For dataset $d$, they are defined as

\begin{equation}
    \mathrm{MAE}_d =
    \frac{1}{N_d}
    \sum_{i=1}^{N_d}
    \left|y_i-\hat{y}_i\right|,
\end{equation}

\begin{equation}
    \mathrm{RMSE}_d =
    \sqrt{
        \frac{1}{N_d}
        \sum_{i=1}^{N_d}
        \left(y_i-\hat{y}_i\right)^2
    },
\end{equation}

where $y_i$ and $\hat{y}_i$ denote the ground-truth and predicted values,
respectively, and $N_d$ is the number of evaluated values in dataset $d$.
The forecasting error reported in the main table is

\begin{equation}
    \mathrm{Error}_d =
    \frac{\mathrm{MAE}_d+\mathrm{RMSE}_d}{2}.
\end{equation}

To aggregate performance across datasets while retaining the convention
that higher values indicate better performance, we define the forecasting
Score as

\begin{equation}
    \mathrm{Score}_{\mathrm{forecast}} =
    \frac{100}{D}
    \sum_{d=1}^{D}
    \frac{1}{1+\mathrm{Error}_d},
\end{equation}

where $D=9$ is the number of datasets. Lower MAE, RMSE, and Error indicate
better forecasting performance, whereas a higher Score is better.

We use Precision@R10 for both supervised classification and unsupervised
anomaly detection. Precision@R10 is treated as a ranking-based evaluation metric rather
than a calibrated decision threshold for deployment. For each model,
we obtain the predicted warning scores for all samples in the test
set and rank the samples in descending order of their scores. We
then compute precision and recall over the ranked predictions and
report the precision at the smallest ranked prefix for which the
recall of the warning class reaches or exceeds 10\%. The validation set is used for model selection and early stopping.

Warning samples are treated as the positive class. For each dataset,
we report the precision at the smallest ranked prefix for which
warning recall reaches or exceeds 10\%:
\begin{equation}
    \mathrm{P@R10}_d
    =
    \mathrm{Precision}_d(k_d^*),
    \qquad
    k_d^*
    =
    \min\{k:\mathrm{Recall}_d(k)\geq 0.10\},
\end{equation}
where $k_d^*$ denotes the corresponding ranked prefix size. The
aggregate Score is the average Precision@R10 across the nine datasets,
expressed on a scale from 0 to 100:

\begin{equation}
    \mathrm{Score}_{\mathrm{cls/ad}} =
    \frac{100}{D}
    \sum_{d=1}^{D}
    \mathrm{P@R10}_d.
\end{equation}

Higher Precision@R10 and Score indicate better performance.

\section{Showcases of Model Performance}

\subsection{Forecasting Showcases}

\begin{figure*}[t]
    \centering
    \includegraphics[width=\textwidth]{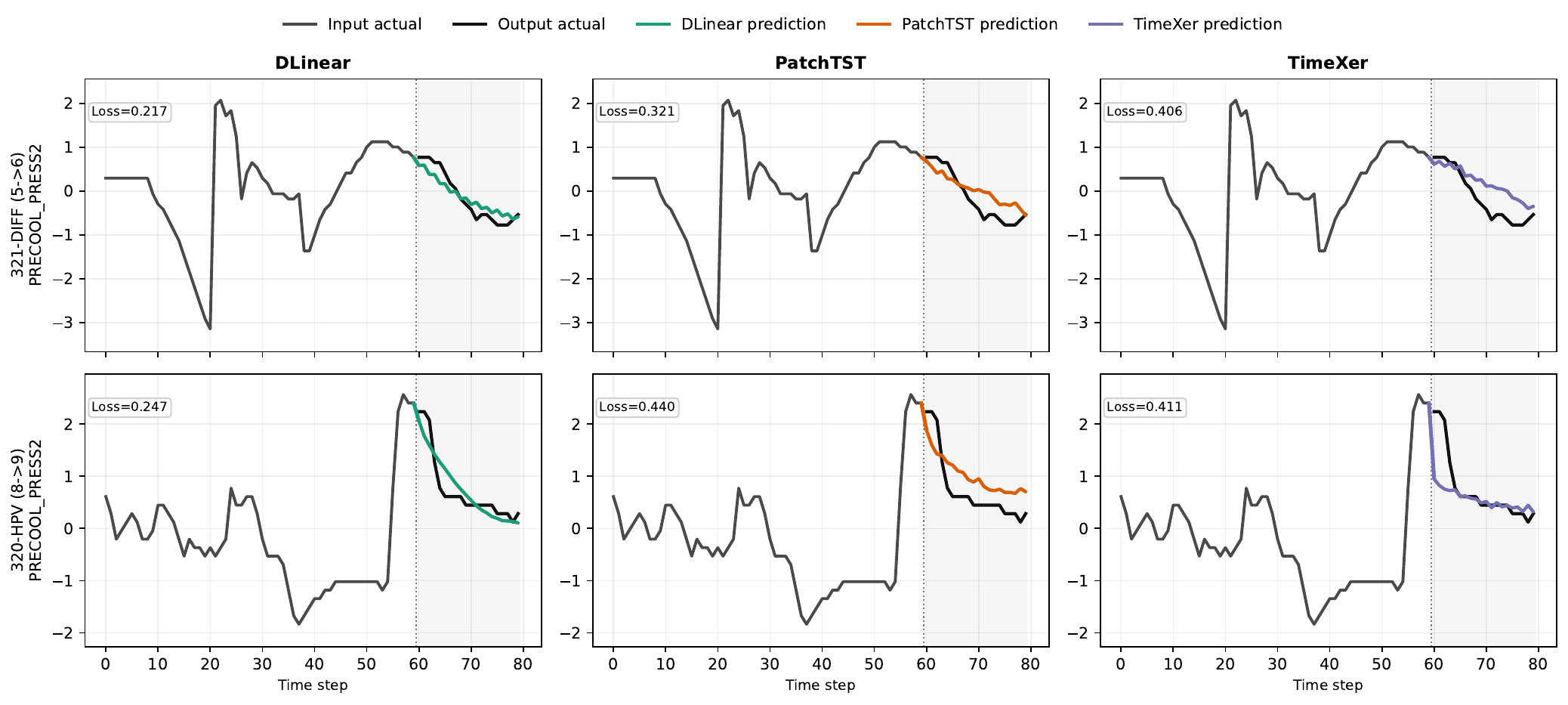}
    \caption{
        Visualization results of DLinear, PatchTST, and
        TimeXer on two representative aircraft time series samples.
    }
    \label{fig:forecasting_showcase}
\end{figure*}

To provide a qualitative comparison beyond the aggregate metrics, we
visualize the forecasting results of DLinear, PatchTST, and TimeXer on
two representative samples from different aircraft subsets and flight
phase transitions. As shown in Figure~\ref{fig:forecasting_showcase},
the historical input and future ground truth are separated by the
vertical dotted line, and the colored curves denote the predictions
produced by the three models.

For the 321-DIFF sample during the transition from Stage 5 (Climbing) to Stage 6 (Cruising), DLinear closely follows the overall decline of the target variable and
achieves the lowest loss of 0.217. PatchTST captures the main direction
of change but exhibits a noticeable level bias, while TimeXer produces
a smoother trajectory and underestimates the subsequent decrease. A
similar pattern is observed for the 320-HPV sample during the transition
from Stage 8 (Cruising) to Stage 9 (Descending). DLinear better reproduces the rapid decrease following the stage transition and attains a loss of 0.247, whereas
PatchTST and TimeXer show larger deviations in the early prediction
region. These examples qualitatively support the quantitative results
reported in the main text, while also illustrating the difficulty of
capturing abrupt state transitions and fine-grained variations in
aircraft operational time series.

\subsection{Classification Showcases}

\begin{figure*}[t]
\centering
\includegraphics[width=0.92\textwidth]{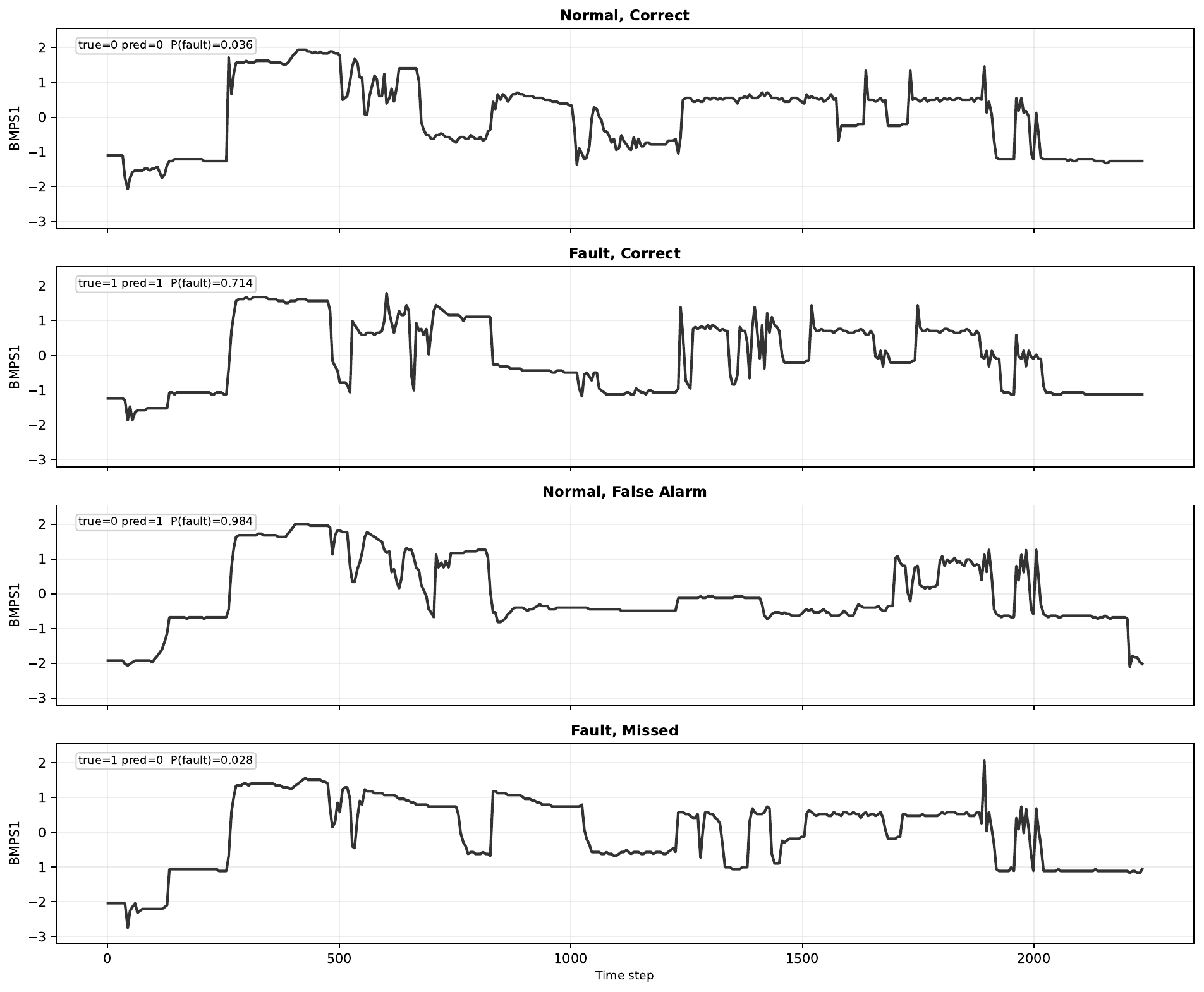}
\caption{
Classification results of DLinear on four representative flight
samples, including a correctly classified normal sample, a correctly
classified fault sample, a false alarm, and a missed fault.
}
\label{fig}
\end{figure*}

To further illustrate the classification behavior, we visualize four representative samples classified by DLinear, covering a correctly identified normal sample, a correctly identified fault sample, a false alarm, and a missed fault. For each sample, the four flight stages are concatenated in temporal order to form a complete flight-level sequence, following the same construction used in the classification experiments. Figure~\ref{fig} presents one representative pressure variable together with the ground-truth label, predicted label, and predicted fault probability.

The correctly classified examples show that the model can distinguish certain fault-related temporal patterns from normal operational behavior. However, the false alarm and missed fault cases reveal substantial overlap between the two classes. In particular, some normal records contain pronounced variations that resemble fault signatures, whereas some faulty records exhibit only weak deviations from normal operating patterns. The high confidence assigned to the false alarm and the low fault probability assigned to the missed sample further indicate that these errors cannot be attributed solely to prediction uncertainty. Instead, they reflect the difficulty of separating subtle fault symptoms from complex operational variation using flight-level time series.

\section{Full Results}

\subsection{Full Results of Temporal Patch Length Analysis}
\label{app:patch_length_results}

Table \ref{tab:patch_length_full_results} reports the complete results of the
temporal patch length analysis presented in
Section \ref{sec:patch_length}. We evaluate PatchTST, TimeXer, and xPatch with
patch lengths of $1$, $2$, $4$, $8$, and $16$, while keeping all other
experimental settings unchanged. The reported loss is computed as the average
of MAE and RMSE and is further averaged across all evaluated subsets and four
flight phase transitions. Lower values indicate better forecasting
performance, and the best result for each model is highlighted in bold.

\begin{table*}[t]
    \centering
    \caption{
        Full forecasting results under different temporal patch lengths,
        broken down by dataset. Loss is averaged across the four flight
        phase transitions. Lower values indicate better performance.
    }
    \label{tab:patch_length_full_results}
    \setlength{\tabcolsep}{5pt}
    \renewcommand{\arraystretch}{1.12}
    \small
    \resizebox{\textwidth}{!}{
    \begin{tabular}{l|*{5}{c}|*{5}{c}|*{5}{c}}
        \toprule
        \multirow{2}{*}{\textbf{Dataset}}
        & \multicolumn{5}{c}{\textbf{PatchTST}}
        & \multicolumn{5}{c}{\textbf{TimeXer}}
        & \multicolumn{5}{c}{\textbf{xPatch}} \\
        \cmidrule(lr){2-6} \cmidrule(lr){7-11} \cmidrule(lr){12-16}
        & \textbf{1} & \textbf{2} & \textbf{4} & \textbf{8} & \textbf{16}
        & \textbf{1} & \textbf{2} & \textbf{4} & \textbf{8} & \textbf{16}
        & \textbf{1} & \textbf{2} & \textbf{4} & \textbf{8} & \textbf{16} \\
        \midrule
        320-DIFF
        & 0.429 & 0.420 & 0.418 & 0.417 & 0.419
        & 0.402 & 0.397 & 0.397 & 0.408 & 0.411
        & 0.409 & 0.407 & 0.407 & 0.404 & 0.403 \\
        320-HPV
        & 0.434 & 0.428 & 0.424 & 0.424 & 0.429
        & 0.409 & 0.410 & 0.408 & 0.416 & 0.417
        & 0.416 & 0.413 & 0.413 & 0.415 & 0.410 \\
        320-PRV
        & 0.430 & 0.426 & 0.424 & 0.421 & 0.424
        & 0.406 & 0.402 & 0.402 & 0.413 & 0.413
        & 0.413 & 0.409 & 0.412 & 0.409 & 0.408 \\
        320-WEAV
        & 0.456 & 0.450 & 0.450 & 0.450 & 0.450
        & 0.443 & 0.437 & 0.436 & 0.451 & 0.455
        & 0.471 & 0.440 & 0.439 & 0.438 & 0.435 \\
        320-LOW
        & 0.458 & 0.446 & 0.446 & 0.443 & 0.446
        & 0.441 & 0.434 & 0.433 & 0.447 & 0.451
        & 0.441 & 0.434 & 0.434 & 0.433 & 0.431 \\
        321-DIFF
        & 0.397 & 0.390 & 0.390 & 0.390 & 0.392
        & 0.378 & 0.378 & 0.375 & 0.386 & 0.387
        & 0.381 & 0.378 & 0.380 & 0.377 & 0.376 \\
        321-WEAV
        & 0.493 & 0.482 & 0.477 & 0.475 & 0.481
        & 0.469 & 0.459 & 0.460 & 0.470 & 0.474
        & 0.466 & 0.462 & 0.462 & 0.460 & 0.459 \\
        787-CAC
        & 0.520 & 0.513 & 0.508 & 0.512 & 0.516
        & 0.469 & 0.455 & 0.454 & 0.462 & 0.466
        & 0.490 & 0.481 & 0.486 & 0.480 & 0.479 \\
        777-PRSOV
        & 0.464 & 0.463 & 0.457 & 0.459 & 0.461
        & 0.461 & 0.451 & 0.452 & 0.464 & 0.464
        & 0.452 & 0.450 & 0.450 & 0.449 & 0.446 \\
        \bottomrule
    \end{tabular}
    }
\end{table*}

\subsection{Full Forecasting Results}
\label{sec:full_forecasting_results}

Due to the space limitation of the main text, we provide the complete
forecasting results in this section. Tables~\ref{tab:detailed_forecasting_mae}
and~\ref{tab:detailed_forecasting_rmse} report the MAE and RMSE results,
respectively, for all evaluated models across the aircraft fault subsets
and flight phase transitions. All experiments follow the same settings
described in the main text. Lower values indicate better forecasting
performance, and the best result in each column is highlighted in bold.
These detailed results complement the aggregated results reported in the
main text and enable a more comprehensive comparison across aircraft
types, fault scenarios, and operating stages.
\begin{table*}[t]
    \centering
    \caption{
        Detailed MAE results across nine datasets and four
        prediction stages. The average result over the four stages is
        reported in the \textit{Avg} row. For each stage the best
        (lowest) result is highlighted in \textbf{bold}, and the
        \textbf{1st Count} row reports the number of stages on which
        each model ranks first. Lower values indicate better performance.
    }
    \label{tab:detailed_forecasting_mae}

    \setlength{\tabcolsep}{3.0pt}
    \renewcommand{\arraystretch}{1.05}
    \scriptsize

    \resizebox{\linewidth}{!}{%
    \begin{tabular}{
        @{}ll|
        *{10}{c}|
        *{5}{c}
        @{}
    }
        \toprule

        \multirow{2}{*}{\textbf{Dataset}}
        & \multirow{2}{*}{\textbf{Stage}}
        & \multicolumn{10}{c|}{
            \textbf{Time Series Forecasting Models (Full-shot)}
        }
        & \multicolumn{5}{c}{
            \textbf{Foundation Models (5-shot)}
        } \\

        \cmidrule(lr){3-12}
        \cmidrule(lr){13-17}

        &
        & \textbf{OLi.}
        & \textbf{xPa.}
        & \textbf{DUET}
        & \textbf{TMix.}
        & \textbf{TXer}
        & \textbf{iTr.}
        & \textbf{DLi.}
        & \textbf{PTST}
        & \textbf{TNet}
        & \textbf{Auto.}
        & \textbf{TiRex}
        & \textbf{Chr.}
        & \textbf{TFM}
        & \textbf{Toto}
        & \textbf{Moirai} \\

        \midrule

        \multirow{5}{*}{320-DIFF}
        & Taxi${\rightarrow}$TakeOff
        & 0.130 & 0.104 & 0.130 & 0.154 & 0.142
        & 0.123 & 0.124 & 0.136 & 0.114 & 0.177
        & 0.107 & \textbf{0.101} & 0.131 & 0.105 & 0.239 \\

        & TakeOff${\rightarrow}$Climb
        & 0.280 & 0.267 & 0.273 & 0.318 & 0.290
        & 0.286 & \textbf{0.180} & 0.284 & 0.277 & 0.389
        & 0.251 & 0.264 & 0.273 & 0.240 & 0.280 \\

        & Climb${\rightarrow}$Cruise
        & 0.287 & 0.278 & 0.288 & 0.304 & 0.310
        & 0.285 & \textbf{0.144} & 0.297 & 0.287 & 0.436
        & 0.272 & 0.276 & 0.280 & 0.263 & 0.303 \\

        & Cruise${\rightarrow}$Descend
        & 0.381 & 0.378 & 0.367 & 0.400 & 0.393
        & 0.384 & \textbf{0.220} & 0.382 & 0.378 & 0.491
        & 0.353 & 0.349 & 0.348 & 0.341 & 0.366 \\

        \rowcolor{gray!12}
        & \textbf{Avg}
        & 0.269 & 0.257 & 0.265 & 0.294 & 0.284
        & 0.269 & 0.167 & 0.275 & 0.264 & 0.373
        & 0.246 & 0.248 & 0.258 & 0.237 & 0.297 \\
        \midrule

        \multirow{5}{*}{320-HPV}
        & Taxi${\rightarrow}$TakeOff
        & 0.115 & \textbf{0.088} & 0.113 & 0.140 & 0.126
        & 0.105 & 0.108 & 0.123 & 0.095 & 0.167
        & 0.096 & 0.089 & 0.119 & 0.093 & 0.227 \\

        & TakeOff${\rightarrow}$Climb
        & 0.282 & 0.269 & 0.274 & 0.314 & 0.292
        & 0.286 & \textbf{0.180} & 0.289 & 0.275 & 0.343
        & 0.257 & 0.269 & 0.280 & 0.245 & 0.286 \\

        & Climb${\rightarrow}$Cruise
        & 0.310 & 0.302 & 0.307 & 0.329 & 0.332
        & 0.308 & \textbf{0.156} & 0.319 & 0.306 & 0.487
        & 0.295 & 0.299 & 0.300 & 0.285 & 0.323 \\

        & Cruise${\rightarrow}$Descend
        & 0.383 & 0.376 & 0.374 & 0.404 & 0.394
        & 0.385 & \textbf{0.221} & 0.389 & 0.377 & 0.468
        & 0.358 & 0.353 & 0.353 & 0.346 & 0.370 \\

        \rowcolor{gray!12}
        & \textbf{Avg}
        & 0.273 & 0.259 & 0.267 & 0.297 & 0.286
        & 0.271 & 0.166 & 0.280 & 0.263 & 0.366
        & 0.251 & 0.253 & 0.263 & 0.242 & 0.302 \\
        \midrule

        \multirow{5}{*}{320-PRV}
        & Taxi${\rightarrow}$TakeOff
        & 0.132 & 0.107 & 0.132 & 0.157 & 0.145
        & 0.126 & 0.125 & 0.141 & 0.114 & 0.188
        & 0.111 & \textbf{0.104} & 0.136 & 0.108 & 0.242 \\

        & TakeOff${\rightarrow}$Climb
        & 0.284 & 0.269 & 0.278 & 0.318 & 0.293
        & 0.289 & \textbf{0.179} & 0.288 & 0.280 & 0.340
        & 0.252 & 0.267 & 0.275 & 0.242 & 0.283 \\

        & Climb${\rightarrow}$Cruise
        & 0.286 & 0.278 & 0.286 & 0.304 & 0.311
        & 0.284 & \textbf{0.143} & 0.297 & 0.286 & 0.463
        & 0.272 & 0.277 & 0.282 & 0.264 & 0.303 \\

        & Cruise${\rightarrow}$Descend
        & 0.382 & 0.375 & 0.371 & 0.399 & 0.391
        & 0.382 & \textbf{0.220} & 0.385 & 0.376 & 0.500
        & 0.352 & 0.349 & 0.348 & 0.340 & 0.365 \\

        \rowcolor{gray!12}
        & \textbf{Avg}
        & 0.271 & 0.258 & 0.267 & 0.294 & 0.285
        & 0.270 & 0.167 & 0.278 & 0.264 & 0.373
        & 0.247 & 0.249 & 0.260 & 0.239 & 0.298 \\
        \midrule

        \multirow{5}{*}{320-WEAV}
        & Taxi${\rightarrow}$TakeOff
        & 0.104 & 0.064 & 0.108 & 0.129 & 0.124
        & 0.077 & 0.083 & 0.101 & 0.108 & 0.165
        & 0.061 & \textbf{0.056} & 0.091 & 0.057 & 0.228 \\

        & TakeOff${\rightarrow}$Climb
        & 0.359 & 0.355 & 0.334 & 0.423 & 0.378
        & 0.367 & \textbf{0.233} & 0.374 & 0.341 & 0.495
        & 0.325 & 0.300 & 0.356 & 0.290 & 0.370 \\

        & Climb${\rightarrow}$Cruise
        & 0.317 & 0.311 & 0.317 & 0.353 & 0.337
        & 0.318 & \textbf{0.210} & 0.318 & 0.321 & 0.455
        & 0.294 & 0.281 & 0.318 & 0.278 & 0.325 \\

        & Cruise${\rightarrow}$Descend
        & 0.386 & 0.374 & 0.380 & 0.423 & 0.415
        & 0.383 & \textbf{0.236} & 0.383 & 0.398 & 0.589
        & 0.336 & 0.323 & 0.350 & 0.317 & 0.369 \\

        \rowcolor{gray!12}
        & \textbf{Avg}
        & 0.291 & 0.276 & 0.285 & 0.332 & 0.313
        & 0.286 & 0.191 & 0.294 & 0.292 & 0.426
        & 0.254 & 0.240 & 0.279 & 0.236 & 0.323 \\
        \midrule

        \multirow{5}{*}{320-LOW}
        & Taxi${\rightarrow}$TakeOff
        & 0.104 & 0.062 & 0.108 & 0.128 & 0.122
        & 0.075 & 0.082 & 0.096 & 0.108 & 0.179
        & 0.060 & \textbf{0.055} & 0.089 & 0.056 & 0.226 \\

        & TakeOff${\rightarrow}$Climb
        & 0.354 & 0.352 & 0.331 & 0.424 & 0.370
        & 0.363 & \textbf{0.232} & 0.370 & 0.331 & 0.471
        & 0.322 & 0.296 & 0.352 & 0.285 & 0.367 \\

        & Climb${\rightarrow}$Cruise
        & 0.317 & 0.312 & 0.317 & 0.352 & 0.337
        & 0.318 & \textbf{0.211} & 0.317 & 0.323 & 0.456
        & 0.296 & 0.283 & 0.320 & 0.281 & 0.329 \\

        & Cruise${\rightarrow}$Descend
        & 0.389 & 0.373 & 0.377 & 0.424 & 0.417
        & 0.383 & \textbf{0.232} & 0.383 & 0.391 & 0.599
        & 0.335 & 0.322 & 0.351 & 0.317 & 0.369 \\

        \rowcolor{gray!12}
        & \textbf{Avg}
        & 0.291 & 0.275 & 0.283 & 0.332 & 0.312
        & 0.285 & 0.189 & 0.291 & 0.288 & 0.426
        & 0.253 & 0.239 & 0.278 & 0.235 & 0.323 \\
        \midrule

        \multirow{5}{*}{321-DIFF}
        & Taxi${\rightarrow}$TakeOff
        & 0.109 & \textbf{0.080} & 0.108 & 0.139 & 0.120
        & 0.095 & 0.102 & 0.114 & 0.084 & 0.121
        & 0.090 & 0.087 & 0.117 & 0.088 & 0.230 \\

        & TakeOff${\rightarrow}$Climb
        & 0.232 & 0.225 & 0.229 & 0.290 & 0.247
        & 0.237 & \textbf{0.150} & 0.240 & 0.234 & 0.358
        & 0.236 & 0.250 & 0.258 & 0.232 & 0.274 \\

        & Climb${\rightarrow}$Cruise
        & 0.289 & 0.283 & 0.284 & 0.305 & 0.311
        & 0.294 & \textbf{0.139} & 0.305 & 0.288 & 0.408
        & 0.281 & 0.281 & 0.289 & 0.270 & 0.307 \\

        & Cruise${\rightarrow}$Descend
        & 0.381 & 0.370 & 0.357 & 0.395 & 0.388
        & 0.379 & \textbf{0.218} & 0.381 & 0.381 & 0.496
        & 0.350 & 0.347 & 0.346 & 0.341 & 0.368 \\

        \rowcolor{gray!12}
        & \textbf{Avg}
        & 0.252 & 0.239 & 0.244 & 0.282 & 0.266
        & 0.251 & 0.152 & 0.260 & 0.247 & 0.346
        & 0.239 & 0.241 & 0.253 & 0.233 & 0.295 \\
        \midrule

        \multirow{5}{*}{321-WEAV}
        & Taxi${\rightarrow}$TakeOff
        & 0.079 & 0.059 & 0.089 & 0.087 & 0.103
        & 0.066 & \textbf{0.045} & 0.102 & 0.070 & 0.137
        & 0.069 & 0.065 & 0.094 & 0.065 & 0.243 \\

        & TakeOff${\rightarrow}$Climb
        & 0.399 & 0.408 & 0.411 & 0.443 & 0.414
        & 0.435 & \textbf{0.253} & 0.440 & 0.408 & 0.468
        & 0.383 & 0.369 & 0.390 & 0.355 & 0.426 \\

        & Climb${\rightarrow}$Cruise
        & 0.312 & 0.307 & 0.291 & 0.327 & 0.321
        & 0.312 & \textbf{0.182} & 0.316 & 0.318 & 0.424
        & 0.303 & 0.298 & 0.324 & 0.289 & 0.342 \\

        & Cruise${\rightarrow}$Descend
        & 0.427 & 0.415 & 0.437 & 0.431 & 0.438
        & 0.432 & \textbf{0.215} & 0.418 & 0.442 & 0.605
        & 0.392 & 0.377 & 0.393 & 0.370 & 0.412 \\

        \rowcolor{gray!12}
        & \textbf{Avg}
        & 0.304 & 0.297 & 0.307 & 0.322 & 0.319
        & 0.311 & 0.174 & 0.319 & 0.309 & 0.408
        & 0.287 & 0.277 & 0.300 & 0.270 & 0.356 \\
        \midrule

        \multirow{5}{*}{787-CAC}
        & Taxi${\rightarrow}$TakeOff
        & 0.343 & 0.333 & 0.328 & 0.379 & 0.338
        & 0.360 & 0.424 & 0.368 & 0.327 & 0.544
        & 0.317 & \textbf{0.272} & 0.331 & 0.287 & 0.370 \\

        & TakeOff${\rightarrow}$Climb
        & 0.308 & 0.349 & 0.321 & 0.419 & 0.322
        & 0.348 & 0.379 & 0.383 & \textbf{0.278} & 0.601
        & 0.354 & 0.318 & 0.387 & 0.331 & 0.418 \\

        & Climb${\rightarrow}$Cruise
        & 0.443 & 0.442 & 0.450 & 0.472 & 0.456
        & 0.443 & 0.462 & 0.450 & 0.451 & 0.721
        & 0.451 & \textbf{0.410} & 0.477 & 0.427 & 0.492 \\

        & Cruise${\rightarrow}$Descend
        & 0.265 & 0.249 & 0.265 & 0.288 & 0.277
        & 0.259 & 0.296 & 0.279 & 0.236 & 0.614
        & 0.244 & \textbf{0.220} & 0.285 & 0.231 & 0.316 \\

        \rowcolor{gray!12}
        & \textbf{Avg}
        & 0.340 & 0.343 & 0.341 & 0.389 & 0.348
        & 0.353 & 0.390 & 0.370 & 0.323 & 0.620
        & 0.342 & 0.305 & 0.370 & 0.319 & 0.399 \\
        \midrule

        \multirow{5}{*}{777-PRSOV}
        & Taxi${\rightarrow}$TakeOff
        & 0.110 & 0.067 & 0.119 & 0.102 & 0.106
        & 0.075 & 0.085 & 0.110 & 0.073 & 0.120
        & 0.082 & \textbf{0.064} & 0.114 & 0.067 & 0.220 \\

        & TakeOff${\rightarrow}$Climb
        & 0.537 & 0.539 & 0.536 & 0.580 & 0.564
        & 0.553 & \textbf{0.376} & 0.555 & 0.533 & 0.653
        & 0.535 & 0.537 & 0.552 & 0.518 & 0.577 \\

        & Climb${\rightarrow}$Cruise
        & 0.205 & 0.197 & 0.201 & 0.235 & 0.225
        & 0.202 & \textbf{0.113} & 0.206 & 0.217 & 0.395
        & 0.205 & 0.201 & 0.232 & 0.194 & 0.271 \\

        & Cruise${\rightarrow}$Descend
        & 0.446 & 0.441 & 0.449 & 0.483 & 0.479
        & 0.449 & \textbf{0.308} & 0.448 & 0.455 & 0.577
        & 0.423 & 0.423 & 0.432 & 0.405 & 0.440 \\

        \rowcolor{gray!12}
        & \textbf{Avg}
        & 0.324 & 0.311 & 0.326 & 0.350 & 0.344
        & 0.320 & 0.221 & 0.330 & 0.320 & 0.436
        & 0.311 & 0.306 & 0.332 & 0.296 & 0.377 \\
        \midrule

        \multicolumn{2}{l|}{\textbf{1st Count}}
        & 0 & 2 & 0 & 0 & 0
        & 0 & \textbf{25} & 0 & 1 & 0
        & 0 & 8 & 0 & 0 & 0 \\

        \bottomrule
    \end{tabular}%
    }

    \vspace{3pt}

    \begin{minipage}{0.98\linewidth}
        \scriptsize
        \textit{Model abbreviations:}
        OLi. = OLinear;
        xPa. = xPatch;
        TMix. = TimeMixer;
        TXer = TimeXer;
        iTr. = iTransformer;
        DLi. = DLinear;
        PTST = PatchTST;
        TNet = TimesNet;
        Auto. = Autoformer;
        TiRex = TiRex-2;
        Chr. = Chronos-2;
        TFM = TimesFM-2.5;
        Toto = Toto-2.0;
        Moirai = Moirai-2.0.
    \end{minipage}
\end{table*}

\begin{table*}[t]
    \centering
    \caption{
        Detailed RMSE results across nine datasets and four
        prediction stages. The average result over the four stages is
        reported in the \textit{Avg} row. For each stage the best
        (lowest) result is highlighted in \textbf{bold}, and the
        \textbf{1st Count} row reports the number of stages on which
        each model ranks first. Lower values indicate better performance.
    }
    \label{tab:detailed_forecasting_rmse}

    \setlength{\tabcolsep}{3.0pt}
    \renewcommand{\arraystretch}{1.05}
    \scriptsize

    \resizebox{\linewidth}{!}{%
    \begin{tabular}{
        @{}ll|
        *{10}{c}|
        *{5}{c}
        @{}
    }
        \toprule

        \multirow{2}{*}{\textbf{Dataset}}
        & \multirow{2}{*}{\textbf{Stage}}
        & \multicolumn{10}{c|}{
            \textbf{Time Series Forecasting Models (Full-shot)}
        }
        & \multicolumn{5}{c}{
            \textbf{Foundation Models (5-shot)}
        } \\

        \cmidrule(lr){3-12}
        \cmidrule(lr){13-17}

        &
        & \textbf{OLi.}
        & \textbf{xPa.}
        & \textbf{DUET}
        & \textbf{TMix.}
        & \textbf{TXer}
        & \textbf{iTr.}
        & \textbf{DLi.}
        & \textbf{PTST}
        & \textbf{TNet}
        & \textbf{Auto.}
        & \textbf{TiRex}
        & \textbf{Chr.}
        & \textbf{TFM}
        & \textbf{Toto}
        & \textbf{Moirai} \\

        \midrule

        \multirow{5}{*}{320-DIFF}
        & Taxi${\rightarrow}$TakeOff
        & 0.351 & 0.378 & 0.332 & 0.449 & 0.378
        & 0.383 & \textbf{0.327} & 0.396 & 0.348 & 0.424
        & 0.379 & 0.371 & 0.424 & 0.378 & 0.535 \\

        & TakeOff${\rightarrow}$Climb
        & 0.534 & 0.533 & 0.518 & 0.583 & 0.541
        & 0.546 & \textbf{0.347} & 0.546 & 0.527 & 0.657
        & 0.523 & 0.534 & 0.540 & 0.511 & 0.536 \\

        & Climb${\rightarrow}$Cruise
        & 0.576 & 0.585 & 0.577 & 0.619 & 0.606
        & 0.590 & \textbf{0.319} & 0.603 & 0.570 & 0.746
        & 0.580 & 0.574 & 0.586 & 0.557 & 0.594 \\

        & Cruise${\rightarrow}$Descend
        & 0.746 & 0.757 & 0.719 & 0.785 & 0.762
        & 0.758 & \textbf{0.470} & 0.759 & 0.732 & 0.829
        & 0.741 & 0.725 & 0.729 & 0.718 & 0.726 \\

        \rowcolor{gray!12}
        & \textbf{Avg}
        & 0.552 & 0.563 & 0.537 & 0.609 & 0.572
        & 0.569 & 0.366 & 0.576 & 0.544 & 0.664
        & 0.556 & 0.551 & 0.570 & 0.541 & 0.598 \\
        \midrule

        \multirow{5}{*}{320-HPV}
        & Taxi${\rightarrow}$TakeOff
        & 0.310 & 0.335 & 0.286 & 0.404 & 0.335
        & 0.337 & \textbf{0.280} & 0.353 & 0.303 & 0.398
        & 0.340 & 0.333 & 0.384 & 0.338 & 0.499 \\

        & TakeOff${\rightarrow}$Climb
        & 0.552 & 0.554 & 0.535 & 0.597 & 0.559
        & 0.564 & \textbf{0.358} & 0.568 & 0.541 & 0.613
        & 0.541 & 0.555 & 0.560 & 0.530 & 0.553 \\

        & Climb${\rightarrow}$Cruise
        & 0.630 & 0.638 & 0.623 & 0.668 & 0.656
        & 0.643 & \textbf{0.353} & 0.655 & 0.624 & 0.816
        & 0.633 & 0.628 & 0.634 & 0.609 & 0.640 \\

        & Cruise${\rightarrow}$Descend
        & 0.749 & 0.761 & 0.722 & 0.789 & 0.762
        & 0.762 & \textbf{0.471} & 0.768 & 0.734 & 0.808
        & 0.749 & 0.731 & 0.738 & 0.727 & 0.734 \\

        \rowcolor{gray!12}
        & \textbf{Avg}
        & 0.560 & 0.572 & 0.542 & 0.614 & 0.578
        & 0.577 & 0.365 & 0.586 & 0.551 & 0.659
        & 0.566 & 0.562 & 0.579 & 0.551 & 0.607 \\
        \midrule

        \multirow{5}{*}{320-PRV}
        & Taxi${\rightarrow}$TakeOff
        & 0.361 & 0.397 & 0.337 & 0.458 & 0.385
        & 0.391 & \textbf{0.327} & 0.409 & 0.353 & 0.455
        & 0.386 & 0.379 & 0.435 & 0.388 & 0.540 \\

        & TakeOff${\rightarrow}$Climb
        & 0.542 & 0.540 & 0.527 & 0.586 & 0.549
        & 0.555 & \textbf{0.351} & 0.554 & 0.534 & 0.605
        & 0.526 & 0.541 & 0.545 & 0.515 & 0.541 \\

        & Climb${\rightarrow}$Cruise
        & 0.577 & 0.585 & 0.573 & 0.617 & 0.607
        & 0.588 & \textbf{0.320} & 0.602 & 0.575 & 0.772
        & 0.581 & 0.576 & 0.589 & 0.559 & 0.594 \\

        & Cruise${\rightarrow}$Descend
        & 0.749 & 0.757 & 0.723 & 0.784 & 0.761
        & 0.758 & \textbf{0.472} & 0.762 & 0.733 & 0.846
        & 0.740 & 0.723 & 0.730 & 0.718 & 0.726 \\

        \rowcolor{gray!12}
        & \textbf{Avg}
        & 0.557 & 0.570 & 0.540 & 0.611 & 0.576
        & 0.573 & 0.367 & 0.582 & 0.549 & 0.669
        & 0.558 & 0.555 & 0.575 & 0.545 & 0.600 \\
        \midrule

        \multirow{5}{*}{320-WEAV}
        & Taxi${\rightarrow}$TakeOff
        & 0.294 & 0.284 & 0.305 & 0.366 & 0.347
        & 0.298 & \textbf{0.273} & 0.294 & 0.337 & 0.401
        & 0.281 & 0.277 & 0.336 & 0.277 & 0.484 \\

        & TakeOff${\rightarrow}$Climb
        & 0.712 & 0.729 & 0.670 & 0.795 & 0.735
        & 0.743 & \textbf{0.473} & 0.756 & 0.699 & 0.868
        & 0.709 & 0.677 & 0.744 & 0.670 & 0.738 \\

        & Climb${\rightarrow}$Cruise
        & 0.658 & 0.665 & 0.647 & 0.699 & 0.676
        & 0.665 & \textbf{0.435} & 0.666 & 0.661 & 0.775
        & 0.649 & 0.632 & 0.675 & 0.632 & 0.663 \\

        & Cruise${\rightarrow}$Descend
        & 0.758 & 0.758 & 0.740 & 0.799 & 0.785
        & 0.762 & \textbf{0.488} & 0.763 & 0.762 & 0.914
        & 0.728 & 0.709 & 0.735 & 0.707 & 0.736 \\

        \rowcolor{gray!12}
        & \textbf{Avg}
        & 0.605 & 0.609 & 0.591 & 0.665 & 0.636
        & 0.617 & 0.417 & 0.620 & 0.615 & 0.739
        & 0.592 & 0.574 & 0.622 & 0.571 & 0.655 \\
        \midrule

        \multirow{5}{*}{320-LOW}
        & Taxi${\rightarrow}$TakeOff
        & 0.289 & 0.276 & 0.299 & 0.361 & 0.342
        & 0.291 & \textbf{0.266} & 0.292 & 0.335 & 0.430
        & 0.275 & 0.272 & 0.327 & 0.271 & 0.478 \\

        & TakeOff${\rightarrow}$Climb
        & 0.700 & 0.719 & 0.659 & 0.790 & 0.721
        & 0.733 & \textbf{0.472} & 0.749 & 0.680 & 0.834
        & 0.702 & 0.672 & 0.736 & 0.659 & 0.733 \\

        & Climb${\rightarrow}$Cruise
        & 0.660 & 0.666 & 0.650 & 0.700 & 0.678
        & 0.666 & \textbf{0.439} & 0.668 & 0.662 & 0.779
        & 0.654 & 0.638 & 0.680 & 0.639 & 0.669 \\

        & Cruise${\rightarrow}$Descend
        & 0.755 & 0.755 & 0.737 & 0.798 & 0.784
        & 0.760 & \textbf{0.481} & 0.761 & 0.759 & 0.920
        & 0.722 & 0.702 & 0.731 & 0.703 & 0.733 \\

        \rowcolor{gray!12}
        & \textbf{Avg}
        & 0.601 & 0.604 & 0.586 & 0.662 & 0.631
        & 0.612 & 0.414 & 0.618 & 0.609 & 0.741
        & 0.588 & 0.571 & 0.619 & 0.568 & 0.653 \\
        \midrule

        \multirow{5}{*}{321-DIFF}
        & Taxi${\rightarrow}$TakeOff
        & 0.275 & 0.289 & \textbf{0.265} & 0.375 & 0.305
        & 0.296 & 0.268 & 0.310 & 0.269 & 0.295
        & 0.311 & 0.303 & 0.364 & 0.305 & 0.491 \\

        & TakeOff${\rightarrow}$Climb
        & 0.486 & 0.489 & 0.476 & 0.548 & 0.502
        & 0.499 & \textbf{0.298} & 0.501 & 0.485 & 0.599
        & 0.508 & 0.527 & 0.527 & 0.501 & 0.546 \\

        & Climb${\rightarrow}$Cruise
        & 0.574 & 0.586 & 0.566 & 0.612 & 0.604
        & 0.595 & \textbf{0.307} & 0.612 & 0.571 & 0.714
        & 0.582 & 0.571 & 0.593 & 0.556 & 0.593 \\

        & Cruise${\rightarrow}$Descend
        & 0.724 & 0.732 & 0.683 & 0.758 & 0.735
        & 0.734 & \textbf{0.463} & 0.737 & 0.721 & 0.851
        & 0.715 & 0.697 & 0.704 & 0.694 & 0.708 \\

        \rowcolor{gray!12}
        & \textbf{Avg}
        & 0.515 & 0.524 & 0.497 & 0.573 & 0.536
        & 0.531 & 0.334 & 0.540 & 0.512 & 0.615
        & 0.529 & 0.524 & 0.547 & 0.514 & 0.585 \\
        \midrule

        \multirow{5}{*}{321-WEAV}
        & Taxi${\rightarrow}$TakeOff
        & 0.262 & 0.252 & 0.268 & 0.288 & 0.282
        & 0.259 & \textbf{0.178} & 0.273 & 0.261 & 0.342
        & 0.275 & 0.269 & 0.298 & 0.264 & 0.479 \\

        & TakeOff${\rightarrow}$Climb
        & 0.771 & 0.788 & 0.767 & 0.829 & 0.783
        & 0.833 & \textbf{0.499} & 0.846 & 0.776 & 0.803
        & 0.774 & 0.751 & 0.770 & 0.735 & 0.796 \\

        & Climb${\rightarrow}$Cruise
        & 0.640 & 0.645 & 0.585 & 0.662 & 0.645
        & 0.645 & \textbf{0.382} & 0.656 & 0.641 & 0.728
        & 0.650 & 0.637 & 0.670 & 0.626 & 0.670 \\

        & Cruise${\rightarrow}$Descend
        & 0.791 & 0.793 & 0.794 & 0.810 & 0.802
        & 0.803 & \textbf{0.437} & 0.795 & 0.803 & 0.920
        & 0.787 & 0.763 & 0.778 & 0.757 & 0.775 \\

        \rowcolor{gray!12}
        & \textbf{Avg}
        & 0.616 & 0.620 & 0.603 & 0.647 & 0.628
        & 0.635 & 0.374 & 0.642 & 0.620 & 0.698
        & 0.622 & 0.605 & 0.629 & 0.596 & 0.680 \\
        \midrule

        \multirow{5}{*}{787-CAC}
        & Taxi${\rightarrow}$TakeOff
        & 0.721 & 0.734 & 0.684 & 0.784 & 0.719
        & 0.792 & 0.833 & 0.794 & 0.703 & 0.881
        & 0.697 & \textbf{0.634} & 0.709 & 0.649 & 0.751 \\

        & TakeOff${\rightarrow}$Climb
        & 0.533 & 0.615 & 0.554 & 0.709 & 0.552
        & 0.619 & 0.654 & 0.675 & \textbf{0.482} & 0.853
        & 0.654 & 0.606 & 0.689 & 0.629 & 0.729 \\

        & Climb${\rightarrow}$Cruise
        & 0.703 & 0.714 & 0.706 & 0.745 & 0.717
        & 0.710 & 0.709 & 0.719 & 0.709 & 0.975
        & 0.714 & \textbf{0.668} & 0.740 & 0.680 & 0.746 \\

        & Cruise${\rightarrow}$Descend
        & 0.494 & 0.499 & 0.495 & 0.541 & 0.511
        & 0.512 & 0.543 & 0.513 & 0.469 & 0.845
        & 0.486 & \textbf{0.448} & 0.527 & 0.466 & 0.545 \\

        \rowcolor{gray!12}
        & \textbf{Avg}
        & 0.613 & 0.640 & 0.610 & 0.695 & 0.625
        & 0.658 & 0.685 & 0.676 & 0.591 & 0.888
        & 0.638 & 0.589 & 0.666 & 0.606 & 0.693 \\
        \midrule

        \multirow{5}{*}{777-PRSOV}
        & Taxi${\rightarrow}$TakeOff
        & 0.179 & \textbf{0.163} & 0.179 & 0.193 & 0.178
        & 0.171 & 0.178 & 0.176 & 0.171 & 0.216
        & 0.215 & \textbf{0.163} & 0.235 & 0.167 & 0.352 \\

        & TakeOff${\rightarrow}$Climb
        & 0.912 & 0.916 & 0.913 & 0.961 & 0.935
        & 0.939 & \textbf{0.648} & 0.934 & 0.903 & 0.995
        & 0.918 & 0.913 & 0.930 & 0.898 & 0.955 \\

        & Climb${\rightarrow}$Cruise
        & 0.417 & 0.423 & 0.423 & 0.473 & 0.451
        & 0.422 & \textbf{0.241} & 0.430 & 0.432 & 0.626
        & 0.433 & 0.418 & 0.453 & 0.412 & 0.489 \\

        & Cruise${\rightarrow}$Descend
        & 0.860 & 0.864 & 0.865 & 0.898 & 0.885
        & 0.869 & \textbf{0.653} & 0.871 & 0.853 & 0.929
        & 0.861 & 0.855 & 0.864 & 0.844 & 0.859 \\

        \rowcolor{gray!12}
        & \textbf{Avg}
        & 0.592 & 0.591 & 0.595 & 0.631 & 0.612
        & 0.600 & 0.430 & 0.603 & 0.590 & 0.691
        & 0.607 & 0.587 & 0.620 & 0.580 & 0.664 \\
        \midrule

        \multicolumn{2}{l|}{\textbf{1st Count}}
        & 0 & 1 & 1 & 0 & 0
        & 0 & \textbf{30} & 0 & 1 & 0
        & 0 & 4 & 0 & 0 & 0 \\

        \bottomrule
    \end{tabular}%
    }

    \vspace{3pt}

    \begin{minipage}{0.98\linewidth}
        \scriptsize
        \textit{Model abbreviations:}
        OLi. = OLinear;
        xPa. = xPatch;
        TMix. = TimeMixer;
        TXer = TimeXer;
        iTr. = iTransformer;
        DLi. = DLinear;
        PTST = PatchTST;
        TNet = TimesNet;
        Auto. = Autoformer;
        TiRex = TiRex-2;
        Chr. = Chronos-2;
        TFM = TimesFM-2.5;
        Toto = Toto-2.0;
        Moirai = Moirai-2.0.
    \end{minipage}
\end{table*}

\end{document}